\documentclass[10pt,twocolumn,letterpaper]{article}

\usepackage{cvpr}              %
\usepackage[accsupp]{axessibility} %

\usepackage[dvipsnames]{xcolor}

\usepackage{multirow}
\usepackage{bbold} %
\usepackage{tabularx} %
\usepackage{colortbl}

\definecolor{cvprblue}{rgb}{0.21,0.49,0.74}
\usepackage[pagebackref,breaklinks,colorlinks,citecolor=cvprblue]{hyperref}

\def\paperID{13577} %
\def\confName{CVPR}
\def\confYear{2024}

\title{CAPE: CAM as a Probabilistic Ensemble for Enhanced DNN Interpretation}

\author{Townim Faisal Chowdhury$^{1}$, 
Kewen Liao$^{2}$, 
Vu Minh Hieu Phan$^{1}$, 
Minh-Son To$^{3}$, 
Yutong Xie$^{1}$, \\
Kevin Hung$^{4}$, 
David Ross$^{4}$, 
Anton van den Hengel$^{1}$, 
Johan W. Verjans$^{1}$, 
\stepcounter{footnote}Zhibin Liao$^{1}$\thanks{Corresponding author.}\\
{\small $^{1}$Australian Institute for Machine Learning, University of Adelaide, Australia}, 
{\small $^{2}$Australian Catholic University, Australia}, \\
{\small $^{3}$Flinders University, Australia},
{\small $^{4}$SA Pathology, Central Adelaide Local Health Network, Australia}\\
}

\begin{document}
\maketitle

\begin{abstract}

Deep Neural Networks (DNNs) are widely used for visual classification tasks, but their complex computation process and black-box nature hinder decision transparency and interpretability. Class activation maps (CAMs) and recent variants provide ways to visually explain the DNN decision-making process by displaying `attention' heatmaps of the DNNs. Nevertheless, the CAM explanation only offers relative attention information, that is, on an attention heatmap, we can interpret which image region is more or less important than the others. However, these regions cannot be meaningfully compared across classes, and the contribution of each region to the model's class prediction is not revealed. To address these challenges that ultimately lead to better DNN Interpretation, in this paper, we propose CAPE, a novel reformulation of CAM that provides a unified and probabilistically meaningful assessment of the contributions of image regions. We quantitatively and qualitatively compare CAPE with state-of-the-art CAM methods on CUB and ImageNet benchmark datasets to demonstrate enhanced interpretability. We also test on a cytology imaging dataset depicting a challenging Chronic Myelomonocytic Leukemia (CMML) diagnosis problem. Code is available at: \textbf{\texttt{\url{https://github.com/AIML-MED/CAPE}}}.
\end{abstract}

\section{Introduction}
Deep neural networks (DNNs), despite achieving superior performance on various tasks such as computer vision and natural language processing, are known to be black boxes~\cite{savage2022breaking} that lack the ability to explain their decision-making process. 
The black-box nature is commonly regarded as a result of the complex model structure characterized by stacked computation layers, involving non-linear functions and many model parameters.
Explainable DNN decisions are crucial to many life-critical scenarios~\cite{van2022explainable} such as AI-powered autonomous driving and medical diagnostics. 
Taking the example of healthcare applications~\cite{caruana2015intelligible}, decision transparency is critical for doctors to understand and trust AI analysis, and to use AI to make insightful and accurate diagnoses or decisions. 

\begin{figure*}[!htbp]
    \centering
    \includegraphics[width=0.8\textwidth]{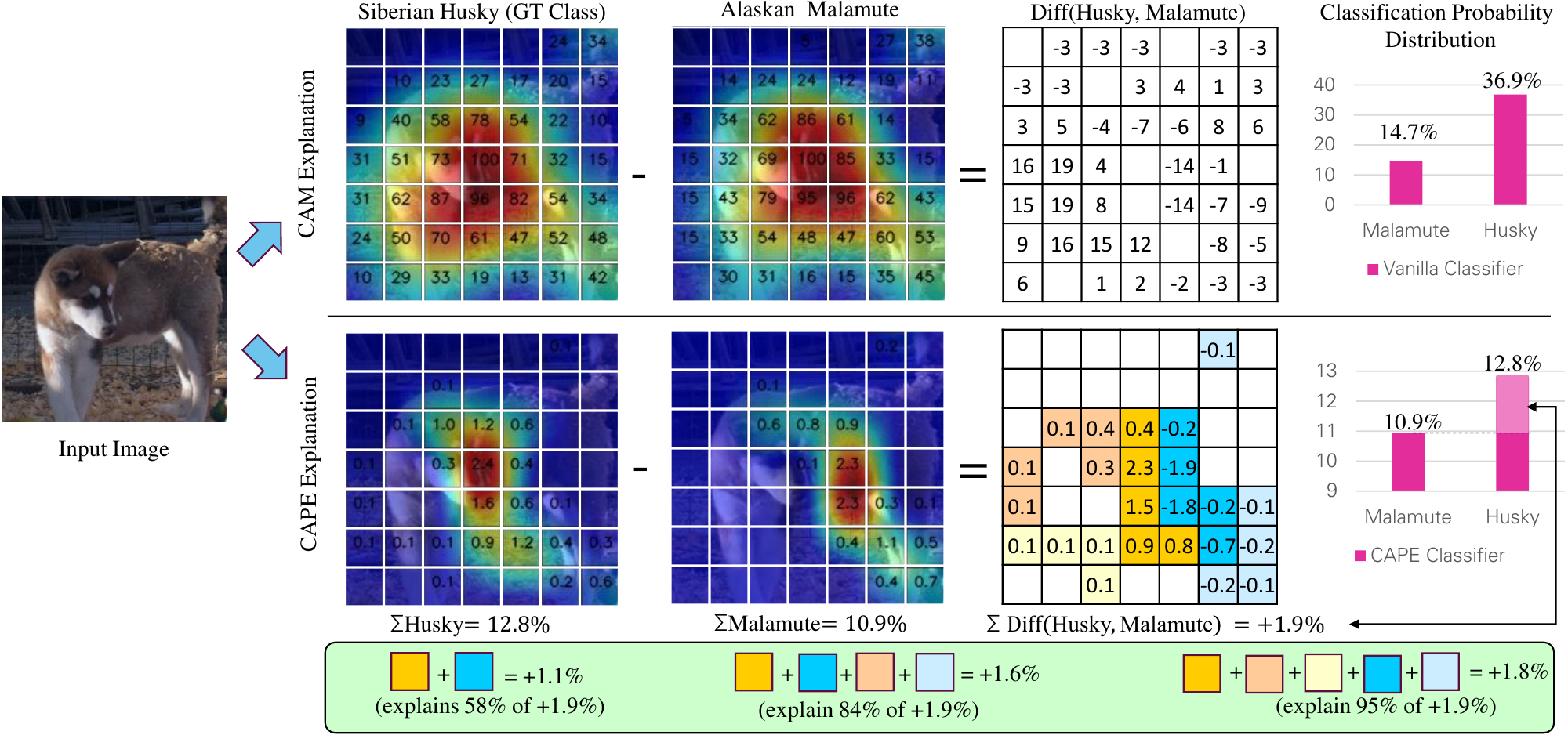}
    \caption{The comparison between CAM and the proposed CAPE explanation methods for a fine-grained class difference analysis example between Siberian Husky (Husky) and Alaskan Malamute (Malamute) classes on ImageNet. 
    We overlay the explanation values before up-sampling on top of the produced heatmaps.
    CAM explanation is class independent which highlights similar regions for similar object classes, making the explanation maps incomparable. Instead, CAPE-produced explanation values (before up-sampling and min-max normalization) are probability values for each spatial location (image region) and class combination. 
    We color code the top-5, next-5 (top-6 to top-10), \etc., for the positive values (\ie, more Husky) and the negative values (\ie, more Malamute) on the Diff graph. 
    The green box shows an example analysis of the $+1.9\%$ class difference by summing the color-coded regions and demonstrating to what levels they explain the class difference. }
    \label{fig:intro}
\end{figure*}

DNN interpretability is an emerging and actively studied research field.
For visual classification tasks, a common type of DNN interpretability analysis is to explain DNN outputs via finding and displaying model attention on the input image, \ie, identifying which image regions the model focused on during the decision-making process. This type of visual explanation can be achieved via methods of gradient-based attention visualization~\cite{simonyan2013deep}, perturbation-based input manipulation~\cite{ribeiro2016should,fong2017interpretable}, and class activation map (CAM)-based visualization~\cite{judd2009learning,selvaraju2017grad}. In particular, CAM is an inherent intermediate step of DNN prediction which represents the actual region relevance produced by the network. CAM stands out due to its efficient feedforward process, yet its attention values can not directly explain and compose model outcomes.
Specifically, CAM values are class-wise relative probabilities. They only represent the relative region importance compared to the highest attention value within each class map. Thus, CAM values provide a limited explanation within the context of one target class. This means that they are incomparable between classes, and cannot explain image-level predictions. 
Take the CAM visualization in Fig.~\ref{fig:intro} as an example, CAM assigns similar attention values to two dog breed classes Siberian Husky and Alaskan Malamute. Differencing the two CAM maps between the breeds fails to yield meaningful comparisons. 

The limited analytical capability of current CAM-based approaches hinders their use in many downstream applications. For example, fine-grained classification analysis requires the model's ability to discriminate regions between closely related concepts.
In addition, for tasks such as weakly supervised semantic segmentation, CAM thresholding is employed to initialize a segmentation method~\cite{kolesnikov2016seed} but the threshold choice is often arbitrary without any semantic meaning attached. 

In this paper, we reformulate \underline{CA}M as a \underline{P}robabilistic \underline{E}nsemble, and name it CAPE. 
Diverging from the current CAM methods, CAPE's activation map seizes the probabilistic and absolute contributions of each image region toward class predictions while enabling meaningful comparisons between classes. As illustrated in Fig.~\ref{fig:intro}, CAPE enforces a direct composition relationship between the overall model prediction and image region contributions. 
Our main contributions are summarized as follows:
\begin{itemize}
    \item We propose a novel CAPE method to explicitly capture the relationship between the model's attention map and the decision process. 
    For each class, the summation of the image region-wise attention values in CAPE is identical to the image-level prediction, providing a basis for the analytical understanding of the model attention.
    \item CAPE inference is efficient, introducing nearly zero extra model parameters and only takes a feed-forward inference to generate the explanation. By reformulating the softmax activation function, CAPE only adds a single trainable scalar, \ie., the Softmax temperature variable.
    \item We discover that CAPE explanation maps tend to highlight class discriminative regions whereas CAM explanation maps are independent for each class that also highlight class mutual regions. Hence, we further propose an alternative class mutual region inclusive CAPE explanation, namely the $\mu$-CAPE ($\mu$ denotes `mutual'), which restores the attention of CAPE on class mutual regions, achieving enhanced performance on commonly evaluated CAM interpretability metrics.
\end{itemize}

\section{Related Work}
In this section, we cover closely related works in interpretable machine learning and softmax-based aggregation.

{\noindent\bf Interpretable Machine Learning.}
For DNNs, the most common interpretation approaches are via saliency/heatmap types of visual explanation using model attention. 
The heatmap visualization methods can be loosely grouped into three categories: gradient-based attention visualization~\cite{simonyan2013deep}, class activation maps (CAMs) base visualization~\cite{judd2009learning,selvaraju2017grad}, and perturbation-based~\cite{ribeiro2016should,fong2017interpretable} input manipulation.
Among them, the CAM method has gained significant research interest due to its ability to produce intuitive and high-quality visual attention~\cite{li2022fd}. 
CAM employs linear weighting of backbone-produced feature maps by using classification layer weights to produce a heatmap for each class category. The heatmap can correspond to class-wise salient regions of the input image. Based on how the CAM's weights are computed, recent works can be categorized into gradient-based and score-based methods. 
Gradient-based CAM methods~\cite{selvaraju2017grad,chattopadhay2018grad,omeiza2019smooth,jiang2021layercam} use the gradients of a target class with respective to the activation maps as a CAM's weights to combine feature maps from the backbone. 
On the other hand, score-based methods such as Score-CAM~\cite{wang2020score} weights CAMs using a score computed by the increase of prediction confidence before and after masking the input image with initial CAM-produced attention. 
A more recent method, FD-CAM~\cite{li2022fd}, leverages gradient-based weights and score-based weights to obtain the CAM's weightings, benefiting from both schemes.
The model-agnostic methods treat models as black boxes that can often be interpreted by input perturbation. 
LIME~\cite{ribeiro2016should} and SHAP~\cite{lundberg2017unified} are two typical model-agnostic methods to explain DNNs via input perturbation. 
They require additional sampling processes and fitting separate explainer models to approximate the original model's inference process, thereby consuming more computations.

{\noindent\bf Softmax-based Aggregation.}
The softmax function gives soft-weighted assignments of member contribution and has the nice property of summing to 1. 
Gao~\etal~\cite{gao2019lip} proposed a softmax-based local importance-based pooling method to down-sample spatial features in receptive fields. The attention mechanism~\cite{bahdanau2014neural} is another example of softmax-based feature aggregation which has been the core component of the modern transformer networks~\cite{vaswani2017attention}.
In capsule networks~\cite{sabour2017dynamic}, softmax is used in the dynamic routing algorithm which can be viewed as a form of parallel attention mechanism to connect capsule layers.
Our proposed interpretation method also utilizes softmax functions to construct probabilistically comparable attention, overcoming CAM's analytical limitation.

\section{Methodology of Model Interpretation}

\subsection{Class Activation Maps (CAMs)}
Let $\mathbf{x}$ be a single image and $y \in \mathcal{C}$ be the corresponding label, where $\mathcal{C}$ denotes the label set of the dataset.
A function $f$ produces a feature tensor from $\mathbf{x}$, \ie, $\mathbf{F} = f(\mathbf{x}; \theta_f)$, where $\mathbf{F} \in \mathbb{R}^{H\times W \times K}$, $H$ and $W$ denote the spatial dimensions and $K$ represents the number of channels.
A typical deep learning classification model utilizes a sequence of a global average pooling layer and a fully-connected layer with a softmax activation function (referred as a vanilla classification layer) to produce the likelihood probability distribution $p(\mathcal{C}|\mathbf{x}, \theta)$ (denote as $\mathbf{p}$) from $\mathbf{F}$, which can be written as:
\begin{equation}
    \mathbf{p} = \mathrm{softmax}_c\bigg(\mathbf{W}^{\intercal}\frac{1}{H \times W} \sum_{i=1}^{H} \sum_{j=1}^{W} \mathbf{F}_{ij} + \mathbf{b}\bigg),   
\label{eq:likelihood}
\end{equation}
where $\theta = \{\theta_f, \mathbf{W} \in \mathbb{R}^{K\times {|\mathcal{C}|}}, \mathbf{b}\in\mathbb{R}^{|\mathcal{C}|}\}$ denotes the set of trainable parameters.
The class activation map $\mathbf{M}_c$ for class $c \in \mathcal{C}$  is obtained by aggregating the activation maps $\mathbf{F}_k$ weighted by their class weights $\mathbf{W}_{kc}$, \ie, $\mathbf{M}_c = \sum_{k=1}^{K}\mathbf{W}_{kc} \mathbf{F}_{k}$ where $\mathbf{M} \in \mathbb{R}^{H\times W\times |\mathcal{C}|}$.
CAM is commonly used as a heatmap type of visualization.
$\mathbf{M}_{ijc}$ indicates the importance of the activation of an image region at position $(i,j)$ toward class $c$. 
For simplicity, we refer an image region as a \textit{pixel} and a 3D indexed element (like $\mathbf{M}_{ijc}$) as a \textit{voxel} hereafter. 
The common approach~\cite{jung2021towards} to produce the explanation (or attention) map $\mathbf{E}$ of the classification model is to apply the rectifier transformation, upsampling, and normalization in sequence: 
\begin{equation}
    \mathbf{E}^{\text{CAM}}_c = \phi\big(\text{max}(\mathbf{M}_c, 0)\big),
\label{eq:ecam}
\end{equation}
where $\phi(.)$ denotes a sequential process of up-sampling and min-max normalization operations.

As shown in the top row of Fig.~\ref{fig:intro}, the normalized CAM explanation map $\mathbf{E}^{\text{CAM}}_{ijc} \in [0, 100\%]$ are not comparable across classes.
Note that the comparability could be restored if the min-max normalization uses the global maximum value of the entire $\mathbf{E}^\text{CAM}$ but even if this is applied, $\mathbf{E}^\text{CAM}$ values only explain the relative importance between voxels but not the absolute importance/contribution toward the model outcome. 
This raises an important research question of \textit{whether CAM methods can show how much each image region actually contributes to the DNN decision}.

As the original CAM formulation ignores the bias term but the bias is involved in model outcome computation, we first restore the bias term by defining shifted CAM maps as $\mathbf{M}' = \mathbf{M} + \mathbf{b}$, and then we define:
\begin{equation}
    p(\mathcal{C}|\mathbf{M}'_{ij}, \mathbf{x}, \theta) = \mathrm{softmax}_c(\mathbf{M}'_{ij}),
\label{eq:contribution}
\end{equation}
to represent the probability distribution of $\mathcal{C}$ at the pixel location $(i,j)$.
Then, a naive way to compute image level prediction $\mathbf{p}$ is to aggregate all pixel probability distributions by averaging:
\begin{equation}
    \hat{\mathbf{p}} = \sum_{i=1}^{H}\sum_{j=1}^{W}\frac{p(\mathcal{C}|\mathbf{M}'_{ij}, \mathbf{x}, \theta)}{H \times W}.
\label{eq:naive_region_contribution}
\end{equation}
Even though $\sum_c\hat{\mathbf{p}} = 1$ appears to satisfy the law of total probability, the model prediction $\mathbf{p}$ and the composed prediction $\hat{\mathbf{p}}$ are not identical, \ie:
\begin{equation}
\mathrm{softmax}_c \big(\frac{1}{H \times W}\sum_{i=1}^{H}\sum_{j=1}^{W}\mathbf{M}'_{ij}\big) \neq\sum_{i=1}^{H}\sum_{j=1}^{W}\frac{\mathrm{softmax}_c(\mathbf{M}'_{ij})}{H \times W},
\label{eq:inequality}    
\end{equation}
because the softmax function is neither additive (\ie, $f(x+y) = f(x)+f(y)$), nor homogeneous (\ie, $f(\alpha x) = \alpha f(x)$).
Therefore, $p(c|\mathbf{M}'_{ij}, \mathbf{x}, \theta) = \frac{\text{softmax}_c(\mathbf{M}'_{ij})}{H \times W}$ is not the true representation of the \textit{voxel contribution to the overall decision} $\mathbf{p}$, and the exact compositional contribution of each voxel to the overall decision $\mathbf{p}$ is intractable.

\begin{figure*}[!h]
    \centering
    \includegraphics[width=0.83\textwidth]{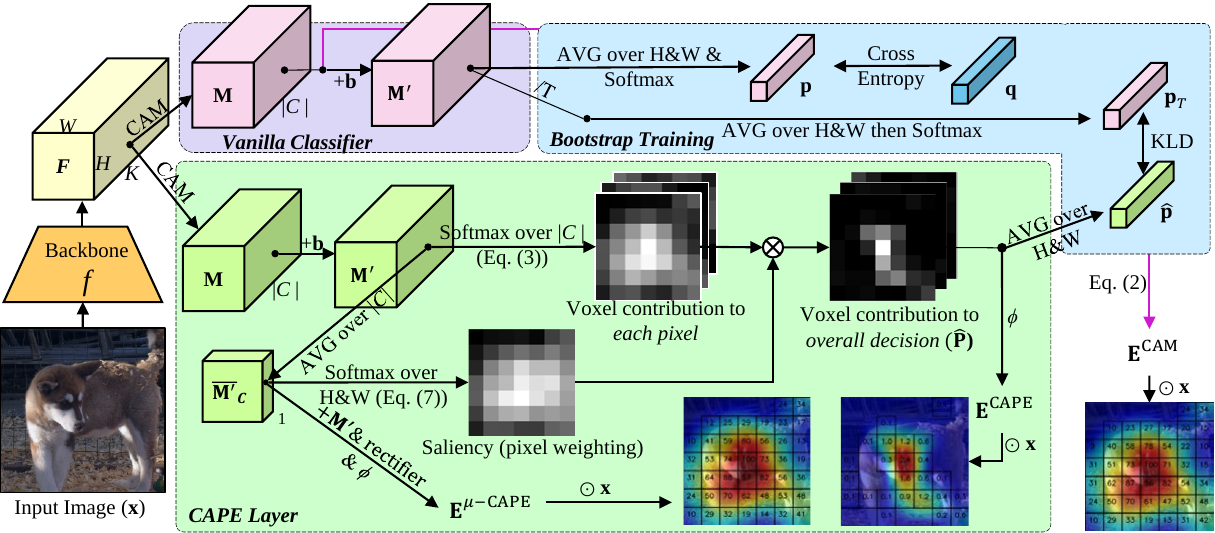}
    \caption{The overview of the proposed CAPE classification layer with bootstrap training. AVG stands for averaging.}
    \label{fig:overview}
\end{figure*}

\subsection{CAM as a Probabilistic Ensemble (CAPE)}

Since the voxel contributions to $\mathbf{p}$ are intractable, we propose to consider $\hat{\mathbf{p}}$ as the model's classification outcome for CAPE.
This allows us to build the relationship between the voxel contributions and the model prediction outcome as a probabilistic ensemble of voxel contributions:
\begin{equation}
    \hat{\mathbf{p}} = \sum_{i=1}^{H} \sum_{j=1}^{W} p(\mathcal{C}|\mathbf{M}'_{ij}, \mathbf{x}, \theta) p(\mathbf{M}'_{ij}|\mathbf{x}, \theta).
\label{eq:CAPE_likelihood}
\end{equation}
The overview of the proposal is depicted in Fig.~\ref{fig:overview} and described as follows.

\subsubsection{Image Region Importance (Saliency)}
From Eq.~(\ref{eq:naive_region_contribution}), we know the naive representation of $p(\mathbf{M}'_{ij}|\mathbf{x}, \theta)$ is $\frac{1}{H \times W}$.
However, to make this formulation less rigid, we apply the softmax aggregation to compute the pixel weighting, also using $\mathbf{M}'$:
\begin{equation}
    p(\mathbf{M}'_{ij}|\mathbf{x}, \theta) = \mathrm{softmax}_{ij}\big(\overline{\mathbf{M}'}_{\mathcal{C}}\big),
\label{eq:weight}
\end{equation}
where the subscripts $ij$ of the softmax function indicate that the softmax normalizes over both spatial dimensions. 
Here, $\overline{\mathbf{M}'}_{\mathcal{C}} = \frac{1}{|\mathcal{C}|}\sum_{c \in \mathcal{C}}\mathbf{M}'_{c}$ denotes the average operation over the classes of $\mathbf{M}'$.
The rationale behind the usage of $\overline{\mathbf{M}'}_{\mathcal{C}}$ comes from the concept of saliency.
When a spatial location processes high activation values in one pixel, it is likely that this pixel contains an object part and therefore should be focused.
The average normalization is used to improve the numerical stability instead of using summation. Note that Eq.~(\ref{eq:weight}) reuses the values from $\mathbf{M}'_c$ which means that the modification to a neural network is only limited to the softmax function in the output layer without introducing additional network parameters.

\subsubsection{CAPE Explanation}
We can compute the \textit{exact} decomposition of overall model prediction $\hat{\mathbf{p}}$ to the contribution from each voxel location $\hat{\mathbf{P}}_{ijc}$ by multiplying Eq.~(\ref{eq:contribution}) and Eq.~(\ref{eq:weight}):
\begin{equation}
    \hat{\mathbf{P}}_{ijc} = \frac{\text{exp}(\mathbf{M}'_{ijc})}{\sum_{c'\in|\mathcal{C}|} \text{exp}(\mathbf{M}'_{ijc'})} \cdot \frac{\text{exp}(\overline{\mathbf{M}'}_{\mathcal{C}})_{ij}}{\sum_{i'j'} \text{exp}(\overline{\mathbf{M}'}_{\mathcal{C}})}_{i'j'}
\label{eq:weighted_contribution}
\end{equation}
where $\hat{\mathbf{P}} \in \mathbb{R}^{H\times W\times |\mathcal{C}|}$.
To form an explanation map, we perform the same operations in Eq.~(\ref{eq:ecam}) and propose: 
$\mathbf{E}^{\text{CAPE}}_c = \phi\big(\hat{\mathbf{P}}_{c}\big)$.
Note $\hat{\mathbf{P}}_{ijc} \in [0, 1]$, therefore clipping negative values of $\hat{\mathbf{P}}_{c}$ is unnecessary. 

\subsubsection{$\mu$-CAPE Explanation}
Although the voxel contributions $\hat{\mathbf{P}}$ are the exact decomposition of the image-level prediction, they do not necessarily produce better quantitative measurement values for the commonly used CAM interpretability evaluation metrics (see CAPE (TS) and (PF) rows in Table~\ref{tab:quantitative_analysis2}).
We found the reason being $\mathbf{E}^{\text{CAPE}}_c$ creates ``sharper'' attention than $\mathbf{E}^{\text{CAM}}_c$ and it tends to place high attention on the class discriminative regions of the objects (\eg, what differentiates Husky and Malamute) but suppresses class mutual regions (\eg, what is common between Husky and Malamute such as dog turso).
We also found that the reason for the sharp attention lies in the super-linearity of the exponential function used in softmax (and when the rectifier function and min-max normalization are applied), which causes the relative distance between the outputs larger than the corresponding inputs, \ie, $\frac{\text{exp}(x) - \text{exp}(y)}{\text{exp}(x) - 0} > \frac{x - y}{x - 0}$, for any $x > y > 0$. 
Therefore, using $\mathbf{E}^{\text{CAPE}}_c \odot \mathbf{x}$ to reclassify the image will take away the decision support from the class mutual regions and cause a large change in the image classification confidence compared to using $\mathbf{E}^{\text{CAM}}_c \odot \mathbf{x}$, resulting lower measurement scores as shown in Table~\ref{tab:quantitative_analysis2}.
A visual illustration can be found in Fig.~\ref{fig:qualitative} between CAM and CAPE (PF) columns.

Intuitively, to restore the class mutual regions, we would retrieve attention values before the softmax normalization, like $\mathbf{M}_{ijc}$ used in CAM.
Therefore, we first transform Eq.~(\ref{eq:weighted_contribution}) to a ``single softmax'' form, taking the advantage of $\text{exp}(x)\text{exp}(y) = \text{exp}(x+y)$:
\begin{equation}
\hat{\mathbf{P}}_{ijc} = \frac{\text{exp}\big(\mathbf{M}'_{ijc} + (\overline{\mathbf{M}'}_{\mathcal{C}})_{ij}\big)}{\sum_{c'\in|\mathcal{C}| }\sum_{i'=1}^{H}\sum_{j'=1}^{W}\text{exp}(\mathbf{M}'_{ijc'} + (\overline{\mathbf{M}'}_{\mathcal{C}})_{i'j'})},
\label{eq:weighted_contribution_new}
\end{equation}
where the CAM equivalent term in CAPE is $\mathbf{M}'_{ijc} + (\overline{\mathbf{M}'}_{\mathcal{C}})_{ij}$.
We define $\mu$-CAPE explanation as
$\mathbf{E}^{\text{$\mu$-CAPE}}_c = \phi\big(\text{max}(\mathbf{M}'_c + \overline{\mathbf{M}'}_{\mathcal{C}}, 0)\big)$.
Note that $\mu$-CAPE restores the class mutual regions but does not maintain the composition relationship to the model outcome.

\subsubsection{Bootstrap Training}
Finally, our loss function is defined in the form of knowledge distillation~\cite{hinton2015distilling}: 
\begin{align}
    \ell = \alpha \cdot \mathcal{H}(\hat{\mathbf{p}}, \mathbf{q}) + \beta \cdot \mathcal{D}_{\text{KL}}(\hat{\mathbf{p}}_{T'}, \mathbf{p}_T),
\label{eq:bootstrap}
\end{align}
where $\mathcal{H}(., .)$ denotes the Cross-Entropy function, $\mathbf{q}$ denotes the classification one-hot label vector, $\text{D}_{\text{KL}}(., .)$ denotes the Kullback–Leibler divergence (KLD) function.
$T'$ denotes the addition of a learnable softmax temperature in Eq.~(\ref{eq:weighted_contribution_new}) and $T$ denotes the addition of a fixed temperature in Eq.~(\ref{eq:likelihood}), both temperature parameters are omitted in the respective equation for clarity.
We propose this form of training using softened $\mathbf{p}_{T}$ as a mediator because the direct optimization using $\mathcal{H}(\hat{\mathbf{p}}, \mathbf{q})$ is difficult, see the classification results of `Direct CE' entry in Table~\ref{tab:classification_performance}.
Once trained, the vanilla classifier could be removed from the CAPE model to maintain nearly identical model parameters except for the learnable temperature parameter.

We further propose two ways of training the CAPE layer. 1) \textbf{training from scratch (TS)} by setting $\alpha=\beta=1$, and during the training, the backbone model does not receive gradients from the CAPE layer but from the vanilla classification layer. 
2) \textbf{post-fitting (PF)} CAPE layer to an already trained classifier model (\eg, ImageNet pre-trained models) by setting $\alpha=0$ and $\beta=1$. 
This means that only the CAPE layer is trained, and we initialize the CAPE layer by the vanilla classifier's parameters. Besides, on the ImageNet dataset, we found the optimization is much more complex as KLD needs to match probability distributions in much higher dimensions.
To alleviate this optimization difficulty, we propose a selective KLD variation that optimizes KLD($\hat{\mathbf{p}}_{T'}, \mathbf{p}_T$) only if the predicted classes of the CAPE layer and the vanilla classification layer are not the same. 
Let $\hat{c} = \text{argmax}_{c'}\hat{\mathbf{p}}_{c'}$ present CAPE predicted class and $c = \text{argmax}_{c'}\mathbf{p}_{c'}$ be the vanilla classifier predicted class,
the selective KLD-enabled bootstrap loss is then:
\begin{equation}
    \ell = \alpha \cdot \mathcal{H}(\mathbf{p}, \mathbf{q}) + \beta \mathbb{1}(\hat{c}\neq c) \cdot \mathcal{D}_{\text{KL}}(\hat{\mathbf{p}}_{T'}, \mathbf{p}_T).
\label{eq:selective_kld}
\end{equation}
The motivation of the design is from the intrinsic prediction discrepancy between CAPE and the vanilla classifier (illustrated in Table~2 in the supplementary material), so it may be unnecessary to match the exact distributions $\hat{\mathbf{p}}_{T'}$ and $\mathbf{p}_T$ for the training samples that $\hat{c} = c$.
With the selective KLD loss, we only bootstrap the prediction distribution once  $\hat{c} \neq c$, which significantly reduces the optimization difficulty. 

\section{Experiments}
\label{sec:exp}

\begin{figure*}[!htbp]
\centering
\includegraphics[width=0.95\textwidth]{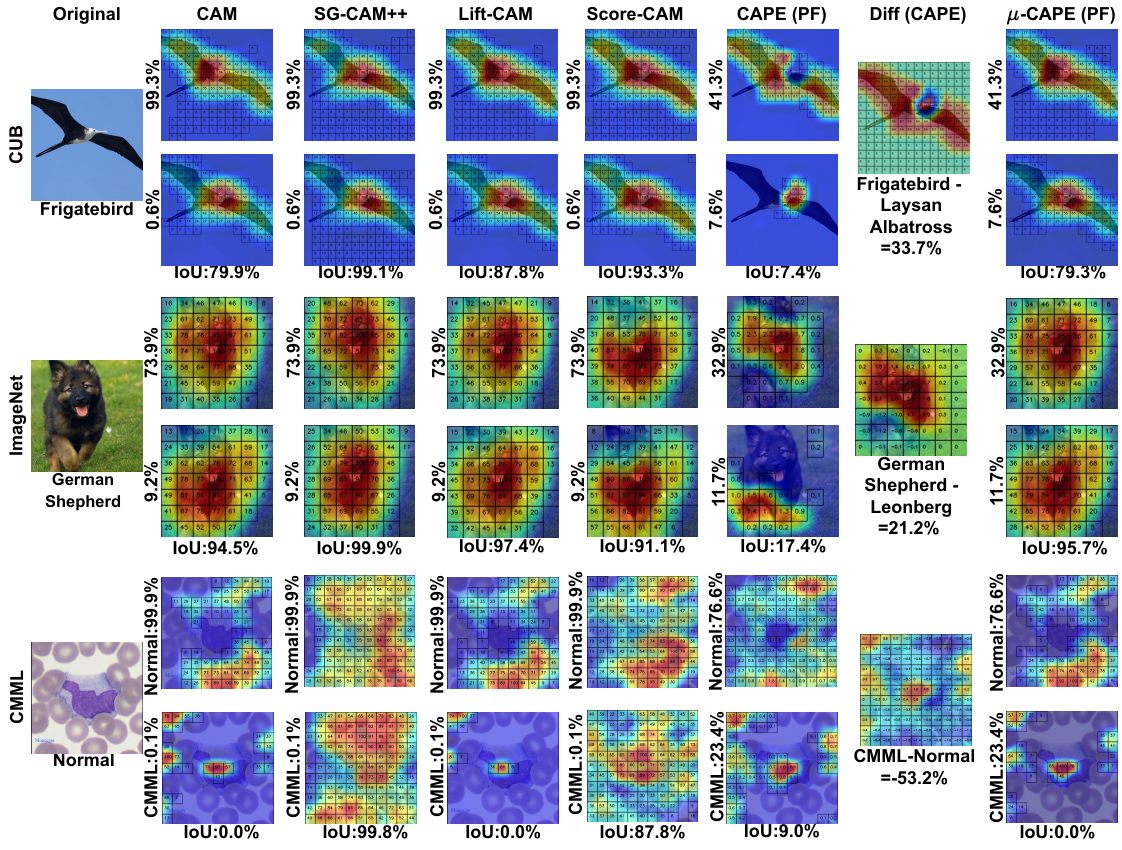}
\caption{Qualitative visualisation using ResNet-50. Each dataset has two rows for the top-2 predicted classes' explanation maps. Class confidence scores are on the left side of each explanation map. We select CAM, Smooth Grad-CAM++, Lift-CAM, and Score-CAM to represent different visualization ways for the same vanilla classification layer. We show CAPE and $\mu$-CAPE (PF) explanations for the proposed CAPE model, full comparisons are in Fig. 2 to 4 in the supplementary material. “SG-CAM++” denotes Smooth Grad-CAM++.
}
\label{fig:qualitative}
\end{figure*}

The proposed CAPE method can be viewed as a replacement for the softmax activation function in the classification module, therefore it is applicable to both DNNs using global average pooling, which can be found in both CNN and Transformer families. 
Therefore, we choose ResNet-50~\cite{he2016deep} and Swin Transformer V2-B~\cite{liu2022swin} as our test beds, please refer to the Section ``Experiments on Swin Transformer model'' in the supplementary material for the results of Swin Transformer model. 
All experiments were conducted on a single Nvidia RTX A6000 GPU (48G video memory) using PyTorch~\cite{NEURIPS2019_9015}. 

\subsection{Datasets and Implementation Details} 
We benchmark on two public datasets: 1) CUB200-2011~\cite{wah2011caltech}; 2) ImageNet ILSVRC2012~\cite{deng2009imagenet}.
We also evaluate a cytology image dataset depicting a difficult Chronic myelomonocytic leukemia (CMML) diagnostic problem. 

\textbf{CUB} comprises a total of 200 distinct bird species, accompanied by 5,995 training images and 5,794 test images. 
The input size is $448\times 448$ and the produced CAM has $14\times14$ spatial size using both ResNet50 and Swin Transformer V2-B model.

\textbf{ImageNet} consists of a total of 1000 object categories with a collection of 1,281,167 images for training and 50,000 images for validation.
We follow the convention in the literature ~\cite{jung2021towards, li2022fd} by randomly selecting 2000 validation images for interpretability evaluation. The input size is $224\times 224$ and the CAM size is $7\times 7$.

\textbf{CMML} dataset contains 3,899 single-cell (Monocyte, a type of white blood cell) images from 171 individuals, who were annotated as `Normal' or having `CMML'. Each individual can have a different amount of monocyte images. 
We report the average results of 5-fold cross-validation on the CMML dataset.
The input size is $352\times 352$ and the derived CAM size is $11 \times 11$.
Additional information on the CMML dataset and motivation for comparing CAMs on CMML can be found in the Section ``CMML Dataset Details" in the supplementary material. 

\textbf{Training Settings.} 
We train all CAPE configurations with SGD optimizer and use temperature $T=2$. 
We set $1e{-4}$ as the initial learning rate for post-fitting (PF) CAPE models and trained them for 30 epochs, except for ImageNet, which is 5 epochs. 
For the training-from-scratch (TS) CAPE models, we uniformly set $1e{-3}$ as the initial learning rate for all three datasets.
We employ step decay with a 0.1 decay rate per 30 epochs and a weight decay of $1e{-4}$ for CUB (200 total epochs) and ImageNet (90 total epochs). 
For the CMML dataset, we use a linear decay with a weight decay of $5e{-4}$ and 100 training epochs by which we reduce the learning rate to $1/100$ of the initial value.
These hyperparameter settings of learning rate and number of epochs follow common settings in the literature. The temperature value was validated on a validation set reserved as a random proportion of the training set. 

\subsection{Qualitative Analysis}
\begin{table*}[!htbp]
\centering
\resizebox{\textwidth}{!}{
\begin{tabular}{c|ccccc|c|ccccc|c|ccccc|c}
\hline
\multirow{2}{*}{Method} & \multicolumn{6}{c|}{CUB} & \multicolumn{6}{c|}{ImageNet} & \multicolumn{6}{c}{CMML} \\
\cline{2-19}
 & AD $\downarrow$ & IC $\uparrow$ & ADD $\uparrow$ & ADCC $\uparrow$ & mIoU $\downarrow$ & BC $\uparrow$ & 
 AD $\downarrow$ & IC $\uparrow$ & ADD $\uparrow$ & ADCC $\uparrow$ & mIoU $\downarrow$ & BC $\uparrow$ & 
 AD $\downarrow$ & IC $\uparrow$ & ADD $\uparrow$ & ADCC $\uparrow$ & mIoU $\downarrow$ & BC $\uparrow$ \\
\hline
CAM & 21.2 & 27.9 & 67.4 & 78.8 & 75.9 & 0 & 
\cellcolor{SpringGreen}\textbf{12.6} & 41.9 & 49.2 & \cellcolor{SpringGreen}\textbf{73.4} & 84.4 & 2 & 
17.4 & 36.0 & \cellcolor{ForestGreen}\textbf{54.8} & \cellcolor{SpringGreen}\textbf{73.6} & \cellcolor{ForestGreen}\textbf{0.1} & \cellcolor{LimeGreen}\textbf{7}\\
Grad-CAM &  21.6 & 27.5 & 66.8 & 77.3 & 100.0 & 0 & 
12.7 & 41.4 & 48.7 & 72.9 & 100.0 & 0 & 
18.2 & 35.3 & \cellcolor{SpringGreen}\textbf{54.0} & 70.6 & 100.0 & 1\\
Grad-CAM++ & 20.3 & 28.7 & 68.9 & 77.4 & 100.0 & 0 & 
13.1 & 39.6 & 47.8 & 72.4 & 100.0 & 0 &
20.1 & 37.7 & 52.5	& 68.4 & 100.0 & 0\\
SG-CAM++ &  23.7 & 24.0 & 64.7 & 74.2 & 99.8 & 0 & 
15.0 & 35.2 & 46.2 & 70.5 & 99.8 & 0 &
31.8 & 31.5 & 47.0	& 70.2 & 99.7 & 0\\
Layer-CAM & 20.1 & 28.7 & 69.9 & 77.3 & 100.0 & 0 & 
13.1 & 39.2 & 48.4 & 71.4 & 100.0 & 0 &
21.6 & 37.1	& 51.8 & 65.9 & 100.0 & 0\\
FD-CAM &  20.5 & 27.9 & \cellcolor{SpringGreen}\textbf{70.9} & 78.1 & 96.7 & 1 &
15.8 & 38.3 & 49.5 & 72.5 & 100.0 & 0 & 
17.8 & 38.9 &	\cellcolor{LimeGreen}\textbf{54.3} &	71.9 & 99.7 & 2\\
LIFT-CAM & 20.9 & 25.6 & 64.5 & 74.6 & 83.3 & 0 &
12.7 & 41.1 & 49.3 & 72.3 & 89.8 & 0 &
\cellcolor{LimeGreen}\textbf{16.4} & 37.8	&	\cellcolor{SpringGreen}\textbf{54.0}	& 72.5  & \cellcolor{ForestGreen}\textbf{0.1} & \cellcolor{SpringGreen}\textbf{6}\\
Score-CAM & \cellcolor{SpringGreen}\textbf{16.3} & \cellcolor{LimeGreen}\textbf{33.0} & \cellcolor{LimeGreen}\textbf{73.1} & \cellcolor{SpringGreen}\textbf{80.2}  & 81.9 & \cellcolor{LimeGreen}\textbf{6} & 
\cellcolor{ForestGreen}\textbf{8.5} & \cellcolor{LimeGreen}\textbf{46.9} & 52.6 & 72.9 & 80.2 & \cellcolor{SpringGreen}\textbf{5} & 
17.0	& \cellcolor{SpringGreen}\textbf{40.3} &	48.3	& 67.7 & 77.0 & 1\\
\hline

CAPE (PF) & 22.2 & 26.5 & 68.7 & 73.7 & \cellcolor{ForestGreen}\textbf{13.4} & 3 & 
17.5 & \cellcolor{SpringGreen}\textbf{45.2} & \cellcolor{ForestGreen}\textbf{59.7} & 69.1 & \cellcolor{ForestGreen}\textbf{11.0} & \cellcolor{LimeGreen}\textbf{7} &
27.9 & 35.0 & 39.9 & 67.9 & 4.9 & 0\\
CAPE (TS) & 27.1 & \cellcolor{SpringGreen}\textbf{31.6} & 59.1 & 77.5 & \cellcolor{LimeGreen}\textbf{28.5} & 3 &
34.7 & 34.2  & 41.3 & 69.9 & \cellcolor{LimeGreen}\textbf{56.8} & 2 & 
29.9	&	27.4	&	36.3	&	72.0 & \cellcolor{SpringGreen}\textbf{0.8} & 1\\
\hline

$\mu$-CAPE (PF) & \cellcolor{LimeGreen}\textbf{15.9} & 30.9 & 69.6 & \cellcolor{LimeGreen}\textbf{83.0} & \cellcolor{SpringGreen}\textbf{66.6} & \cellcolor{SpringGreen}\textbf{5} &
12.7 & 43.9 & \cellcolor{SpringGreen}\textbf{55.9} & \cellcolor{ForestGreen}\textbf{74.3} & \cellcolor{SpringGreen}\textbf{70.3} & \cellcolor{SpringGreen}\textbf{5} & 
\cellcolor{SpringGreen}\textbf{16.5} & \cellcolor{LimeGreen}\textbf{43.6} & 45.5 & \cellcolor{LimeGreen}\textbf{78.2} & \cellcolor{LimeGreen}\textbf{0.6} & \cellcolor{LimeGreen}\textbf{7}\\
$\mu$-CAPE (TS) & \cellcolor{ForestGreen}\textbf{10.3} & \cellcolor{ForestGreen}\textbf{48.5} & \cellcolor{ForestGreen}\textbf{74.2} & \cellcolor{ForestGreen}\textbf{84.4} & 80.9 & \cellcolor{ForestGreen}\textbf{12} &
\cellcolor{LimeGreen}\textbf{10.7} & \cellcolor{ForestGreen}\textbf{58.3} & \cellcolor{LimeGreen}\textbf{58.7} & \cellcolor{LimeGreen}\textbf{73.5} & 89.0 & \cellcolor{ForestGreen}\textbf{9} & 
\cellcolor{ForestGreen}\textbf{14.1}	&	\cellcolor{ForestGreen}\textbf{48.0}	&	50.1	& \cellcolor{ForestGreen}\textbf{78.4} & 5.3 & \cellcolor{ForestGreen}\textbf{9}\\
\hline

\end{tabular}
}
\caption{Comparison of CAM interpretation methods using ResNet-50 backbone model. $\downarrow$ and $\uparrow$ indicate lower or higher is better. ``SG-CAM++'' denotes Smooth Grad-CAM++. The top-3 scores are marked from darker to lighter green colors.}
\label{tab:quantitative_analysis2}
\end{table*}

We compare to eight state-of-the-art DNN-based CAM interpretation methods, including activation-based CAM methods: CAM~\cite{zhou2016learning}, Layer-CAM~\cite{jiang2021layercam}, Score-CAM~\cite{wang2020score}, LIFT-CAM~\cite{jung2021towards}, FD-CAM~\cite{li2022fd}), and gradient-based CAM methods: Grad-CAM~\cite{selvaraju2017grad}, Grad-CAM++~\cite{chattopadhay2018grad}, Smooth Grad-CAM++~\cite{omeiza2019smooth}). 

The qualitative analysis is visualized in Fig.~\ref{fig:qualitative} for CUB, ImageNet, and CMML datasets using the ResNet50 model. 
We compare CAM, Smooth Grad-CAM++, LIFT-CAM, and Score-CAM with our proposed CAPE and $\mu$-CAPE explanations (PF-trained models). 
For each compared method except CAPE, we plot explanation heatmaps for the top-2 predicted classes in the background and overlay their pre-upsampling attention values in the foreground.
Some image region boxes are omitted from the drawing to avoid cluttering and it is based on whether the region's attention value exceeds the 5\% threshold of maximum attention values.
For the compared CAM methods this 5\% threshold is not meaningful.
However, for CAPE visualization, and using the ImageNet ``German shepherd'' dog example, the threshold translates to a minimum probability where below the probability is considered as noise, \ie, $5\%\times2.9\%$ $\approx$ 0.145\% (2.9\% is the largest attention probability), and the kept regions constitute 32.8\% of the class prediction of 32.9\%. 
We can analytically say the regions cropped above the threshold maintain 99.7\% of the original class confidence.

With CAPE's probabilistic ensemble formulation, we can directly compare the two class maps shown in the Diff(CAPE) column.
Furthermore, for the CMML task, predicting CMML from monocyte images is an exploratory and open-ended research question, hence we are particularly interested in understanding where the classifier looks at when the decision is made (\eg, nucleus, cytoplasm, cell exterior region, or their touching boundaries).
Each image's attention placement can be significantly different and hard to manually review beyond a few.
Using CAPE with the additional help of image segmentation, we can easily compute an empirical summary of the attention placement over all test images such as shown in Table~6 of the supplementary material.

CAPE and $\mu$-CAPE are for different purposes.
CAPE is analytical and can explain class discriminative regions which generally show less overlap between the top-2 class explanation maps.
In addition, CAPE achieves lower intersection over union (IoU, defined in Sec.~\ref{sec:quantitative_analysis}). 
These characteristics make CAPE useful for understanding the subtle differences between visually similar concepts.
$\mu$-CAPE shows significant overlap between the top-2 classes and is more useful when the full class object is needed.
An exception is in the CMML example where CAM, Lift-CAM, and $\mu$-CAPE already show the class discriminative characteristic and have 0.0\% IoU, meaning that the respective explanation maps do not overlap. 
This is likely because of the two classes in the CMML problem because the small number of classes trained classifiers are more likely to discard non-discriminative regions~\cite{chen2023extracting}.

\subsection{Quantitative Analysis}
\label{sec:quantitative_analysis}
We use four common CAM interpretability evaluation metrics with an additional metric in our quantitative analysis.
Let $\mathbf{E}_c = \Phi(\mathbf{x}, c)$ denote the overall process that generates an explanation map $\mathbf{E}_c$ from an image given class $c$, and $\mathbf{p}_c = \Psi(\mathbf{x}, c)$ denotes the model prediction generation process. The measurements are defined below. 

\textbf{Average Drop in Confidence (AD)}~\cite{chattopadhay2018grad}. For a single image with target class $c$, $\text{AD}(\mathbf{x}) = \frac{\text{max}(y_c - o_c, 0)}{y_c}$;  where $y_c = \Psi(\mathbf{x}, c)$ and $o_c = \Psi(\mathbf{E}_c \odot \mathbf{x}, c)$, $\odot$ defines the element-wise production, and $c = \text{argmax}_{c'\in |\mathcal{C}|}(\mathbf{p}_{c'})$.

\textbf{Average Increase in Confidence (IC)}~\cite{chattopadhay2018grad} measures the confidence gain when the explanation map is applied: $\text{IC}(\mathbf{x}) = \mathbb{1}(y_c < o_c)$, where $\mathbb{1}$ is an indicator function.

\textbf{AD in Deletion (ADD)}~\cite{jung2021towards} overcomes the drawbacks that IC and AD give good scores when an interpretation method always gives an over-confident explanation. $\text{ADD}(\mathbf{x}) = \frac{\text{max}(y_c - d_c, 0)}{y_c}$, where $d_c = \Psi\big((1 - \mathbf{E}_c) \odot \mathbf{x}, c\big)$.

\textbf{AD, Coherency, and Complexity (ADCC)}~\cite{poppi2021revisiting} was introduced as a robust measurement in comparison to AD and IC.
ADCC represents the harmonic mean of different metrics. 
$\text{ADCC}(\mathbf{x}) = \frac{3}{\text{coh}(\mathbf{E}_c, \mathbf{E}'_c)^{-1} + (1-\text{com}(\mathbf{E}_c))^{-1} + (1-\text{AD}(\mathbf{x}))^{-1}}$.
$\text{coh}(\mathbf{E}_c, \mathbf{E}'_c) = 2\cdot\text{corr}(\mathbf{E}_c, \mathbf{E}'_c) + 1$ measures the min-max normalized Pearson Correlation Coefficient (corr) between $\mathbf{E}_c$ and $\mathbf{E}'_c = \Phi(E_c \odot \mathbf{x, c})$.
$\text{com}(\mathbf{E}_c) = |\mathbf{E}_c|$ measures the complexity of an explanation map by its $\mathcal{L}_1$-norm.

\textbf{Intersection over Union (IoU)} measures the overlap between the explanation maps of top-2 predicted classes. 
We first create a mask $\mathbf{S}_{c} = \mathbf{E}_{c} > 0.2 \cdot \text{max}(\mathbf{E}_{c})$, then $\text{IoU}(\mathbf{x}) = \frac{{|\mathbf{S}_{c_1} \cap \mathbf{S}_{c_2}|}}{{|\mathbf{S}_{c_1} \cup \mathbf{S}_{c_2}|}}$, for the top-2 classes $c_1$ and $c_2$. 
We report the mean IoU (mIoU) in Table~\ref{tab:quantitative_analysis2}.  

\textbf{Borda Count (BC)} is a voting method to give a score based on multiple rankings.
We assign a 1st ranking a score of 3, a 2nd ranking a score of 2, and a 3rd ranking a score of 1.
The rest ranks are scored 0.
Our BC ranking sums over the scores from the above measurements.

The quantitative analysis is shown in Table~\ref{tab:quantitative_analysis2}. We show the following observations.
\begin{enumerate}
    \item $\mu$-CAPE explanations hold the top BC rankings across all datasets because of their AD, IC, ADD, and ADCC scores.
    This illustrates the advantage of the $\mu$-CAPE explanation in terms of the capability to include both class discriminative and class mutual regions. 
    In contrast, the CAPE explanation highlights class discriminative regions hence leading the mIoU measurement.
    \item All $\mu$-CAPE (TS) measurements are generally better than the (PF) model measurements but the PF models are much cheaper to run, especially on large datasets. 
    Comparing CAPE (TS) and (PF), the (PF) version leads to better mIoU on the CUB and ImageNet datasets, but the opposite is observed on the CMML dataset.
    \item Score-CAM has a good BC ranking based on high AD, IC, ADD, and ADCC rankings on CUB and ImageNet.
    Notably, it has a significantly lower AD score on ImageNet. Score-CAM explanation map for an image and a target class pair requires the computation of explanation maps for all classes, which is computationally intensive. In contrast, $\mu$-CAPE and CAPE only need a simple feed-forward inference that incurs trivial computation overhead compared to the original CAM. For instance, CAPE and CAM take around 150 milliseconds to compute for one CUB image on our hardware, and Score-CAM takes 15 seconds.
\end{enumerate}

\subsection{Ablation Study on Classification Performance}
In Table~\ref{tab:classification_performance}, we show the classification performance using the vanilla classification layer and the CAPE classification layer on the same ResNet-50 model with different settings. 
The Naive AVG and Off-the-shelf CAPEs reuse the vanilla classification layer's parameters where their difference is that Naive AVG CAPE aggregates all pixel probability distributions by averaging (Eq.~(\ref{eq:naive_region_contribution})) while Off-the-shelf CAPE employs the image region importance (Eq.~(\ref{eq:CAPE_likelihood})).  
Both models can be used as post hoc \textit{visual} interpretation methods like CAMs, but they have classification performance gaps toward the vanilla classifier.
This leads to our proposal of training the CAPE model to mitigate the gap.
The Direct CE CAPE (TS) employs full course training using the cross-entropy loss $\mathcal{H}(\hat{\mathbf{p}}, \mathbf{q})$ but does not show a significant improvement from Off-the-shelf CAPE.
Both Bootstrap-trained (TS and PF) models get closer performance to the vanilla classification model but arguably there is a marginal performance gap. 
Finally, we stress that our $\mu$-CAPE and CAPE explanations share the same model and only differ in their explanation map formation (\ie., $\mathbf{E}^{\text{$\mu$-CAPE}}_c$ \vs. $\mathbf{E}^{\text{CAPE}}_c$). 

\begin{table}[!htbp]
\centering
\resizebox{0.75\columnwidth}{!}{
\begin{tabular}{c|c|c|c|c}
\hline
\multicolumn{2}{c|}{Model} & CMML & CUB & ImageNet \\
\hline
\multicolumn{2}{c|}{\# Classes ($|\mathcal{C}|$)} & 2 & 200 & 1,000 \\ 
\hline
\multicolumn{2}{c|}{$H\times W$} & $11\times 11$ & $14\times 14$ & $7\times 7$  \\
\hline
\multicolumn{2}{c|}{$H\cdot W \cdot |\mathcal{C}|$} & 242 & 39,200 & 49,000  \\
\hline
\multirow{5}{*}{\rotatebox[origin=c]{90}{CAPE}} & Naive AVG & 89.5 & 79.01 & 74.01\\
& Off-the-shelf & 87.4 & 80.62 & 74.01\\
& Direct CE (TS) & 88.8 & 80.51 & 72.95 \\
& Bootstrap (PF) & 90.3 & 82.12 & 74.42 \\
& Bootstrap (TS) & 89.8 & 82.19 & 74.64 \\
\hline
\multicolumn{2}{c|}{Vanilla classification} & 90.5 & 83.34 & 76.13 \\
\hline
\end{tabular}
}
\caption{Classification accuracy evaluated on ResNet50 model for different CAPE configurations and vanilla classification layer.
}
\label{tab:classification_performance}
\end{table}

\section{Discussion and Conclusion}

We proposed CAPE, a novel DNN interpretation method that is powerful in visualizing and analyzing DNN model attention. It enables us to probabilistically understand how the model predicts, and provides novel insights into meaningful and analytical interpretations.
CAPE is a simple reformulation of the softmax classification layer that adds a trivial cost to classification inference and visual explanation compared to the vanilla classifier and CAM explanation. We conclude with CAPE's characteristics and limitations to motivate future work.

\begin{figure}[!tbp]
    \centering
    \includegraphics[width=0.85\columnwidth]{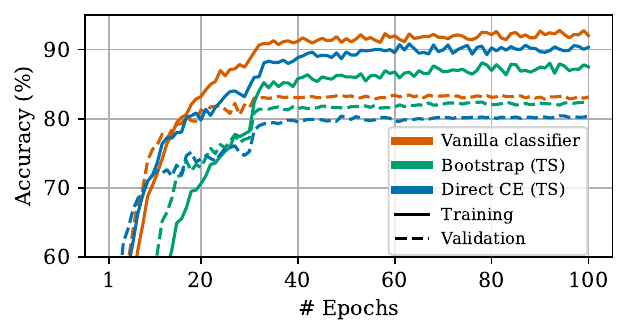}
    \caption{The ResNet-50 training and validation classification accuracy recorded during the training course for the CUB dataset.}
    \label{fig:convergence}
\end{figure}

\textbf{Training convergence and soft prediction confidence.}
Fig.~\ref{fig:convergence} illustrates that the training convergence issue affects the CAPE model's accuracy. 
We believe the convergence issue is caused by the soft prediction confidence characteristic of CAPE (see Table~1 of the supplementary material).
We suspect that the softened predictions in the CAPE formulation are a result of the large number ($H \times W \times |\mathcal{C}|$) of voxels accumulated in the denominator of the softmax function (see Eq.~(\ref{eq:weighted_contribution_new})).
It is commonly known that interpretable models often have to trade accuracy for improved explainability~\cite{panousis2023sparse, yuksekgonul2023posthoc}. 
We believe for CAPE, the trade-off is between the probability computation capacity (leading to improved explainability and analytical ability) and the soft prediction confidence (causing training convergence issues), both resulting from the usage of 
softmax normalization.  
Bootstrap training was introduced to soften the classification confidence scores of the vanilla classifier and therefore mitigate the optimization difficulty of CAPE training.

\textbf{CAPE explains itself.}
Even though the CAPE module's training was bootstrapped from the vanilla classifier and the CAPE models' classification performance approaches to the vanilla classifier's performance, Table~2 in the supplementary material shows an in-negligible prediction disagreement between the two classification layers. 
Hence, CAPE's probabilistic explanation should not be used to explain the decision process of the vanilla classification classifier.

{
    \small
    \bibliographystyle{ieeenat_fullname}
    \bibliography{main}
}

\end{document}


\maketitle              %

\section{Additional Information on ResNet-50}

In this section, we show the tables mentioned by the two discussion points in the section ``Discussion and Conclusion'' of the main manuscript.

\textbf{Prediction Confidence.} 
In Table~\ref{tab:prediction_confidence} we show the prediction confidence on training and validation sets, which demonstrates the need for softened classification prediction ($T=2$) from the vanilla classification layer to guide the training of the CAPE model.
In practice, we found that a larger $T$, \eg, $T=4$ will over-soften the vanilla model's prediction and reduce the performance of trained CAPE.

\begin{table}[!htbp]
\centering
\resizebox{\columnwidth}{!}{
\begin{tabular}{c|c|c|c|c}
\hline
& Classifier Module& CMML & CUB & ImageNet \\
\hline
\multirow{4}{*}{\rotatebox[origin=c]{90}{Train}} & Vanilla classification ($T=1$) & 98.4 & 85.8 & 76.3 \\
& Vanilla classification ($T=2$) & 95.9 & 45.4 & 44.7 \\
& Bootstrap (PF) & 81.9 & 26.7 & 43.8 \\
& Bootstrap (TS) & 88.3 & 19.7 & 6.4 \\
\hline
\multirow{4}{*}{\rotatebox[origin=c]{90}{Val}} & Vanilla classification ($T=1$) & 96.9 & 75.7 & 79.7 \\
& Vanilla classification ($T=2$) & 92.6 & 33.8 & 49.3 \\
& Bootstrap (PF) & 78.5 & 23.5 & 46.7 \\
& Bootstrap (TS) & 84.6 & 17.7 & 6.8 \\
\hline
\end{tabular}
}
\caption{The empirical mean of the prediction confidence over all three reported datasets on the ResNet-50 model.}
\label{tab:prediction_confidence}
\end{table}

\textbf{Prediction Agreement.} 
In Table~\ref{tab:prediction_agreement}, we show prediction agreement between the evaluated models. This demonstrates that each model's explanation is unique and cannot be used to explain each other even if they share the exact model parameters, \ie, the Off-the-shelf model \vs Vanilla Classification model.

\begin{table}[!htbp]
\centering
\resizebox{\columnwidth}{!}{
\begin{tabular}{c|c|c|c|c}
\hline
\multicolumn{2}{c|}{Compared models} & CMML & CUB & ImageNet \\
\hline
Off-the-shelf & Vanilla Classification & 90.5 & 89.0 & 87.4\\
Bootstrap (PF) & Off-the-shelf & 91.9 & 94.9 & 93.8\\
Bootstrap (PF) & Vanilla Classification & 96.6 & 89.3 & 88.1\\
Bootstrap (PF) & Softened Bootstrap (PF) & 98.1 & 99.6 & 94.6\\
Bootstrap (PF) & Bootstrap (TS) & 95.8 & 92.1 & 79.8\\
Bootstrap (TS) & Vanilla Classification & 97.8 & 88.5 & 88.5\\
\hline
\end{tabular}
}
\caption{Prediction agreement (\%) between CAPE (TS/PF) and the Vanilla classification model evaluated on the ResNet-50 model.
Softened Bootstrap (PF) denotes the prediction made by the CAPE layer with learned $T'$ (see Eq.~(10) in the main manuscript).}
\label{tab:prediction_agreement}
\end{table}

\subsection{Additional Qualitative Figures}

\begin{figure*}[!bp]
\centering
\resizebox{1\textwidth}{!}{
\setkeys{Gin}{width=1.1\linewidth}
\newcolumntype{C}{>{\centering\arraybackslash}X}
\begin{tabularx}{\textwidth}{CCCCCCC}
Original & CAM & Grad-CAM & G-CAM++ & SG-CAM++ & CAPE (PF) & $\mu$-CAPE (PF) \\
\multirow{5}{*}{\raisebox{-0.5\height}{\shortstack{\includegraphics{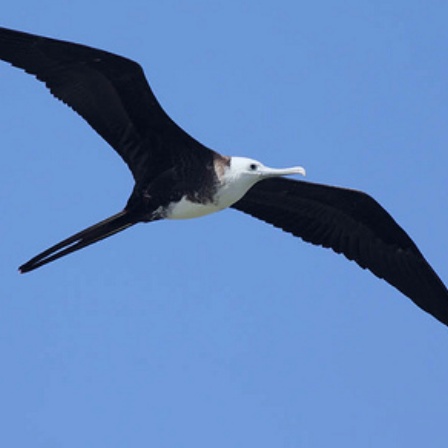}\\Frigatebird}}} & 
\raisebox{-0.5\height}{\includegraphics{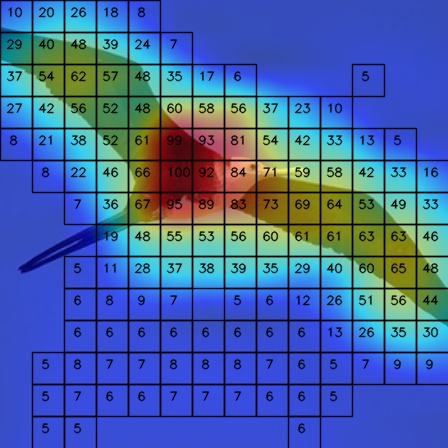}} & 
\raisebox{-0.5\height}{\includegraphics{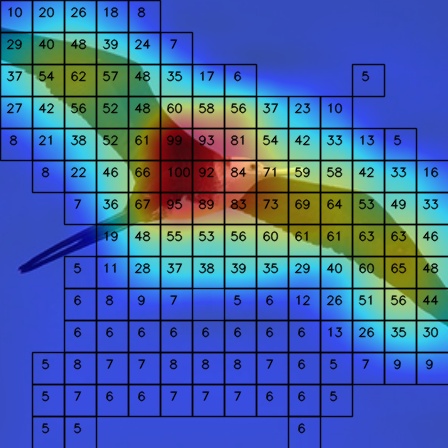}} & 
\raisebox{-0.5\height}{\includegraphics{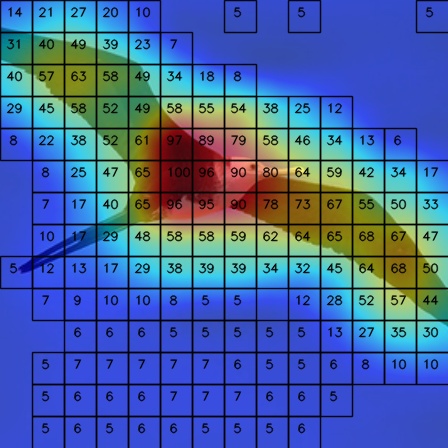}} & 
\raisebox{-0.5\height}{\includegraphics{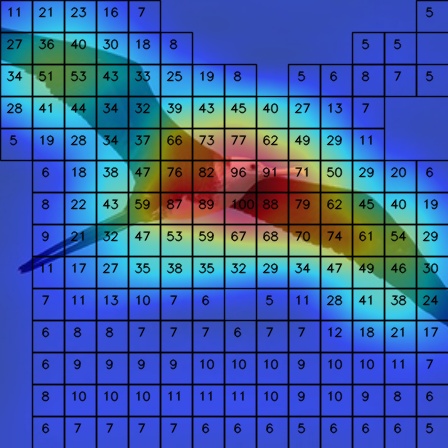}} & 
\raisebox{-0.5\height}{\includegraphics{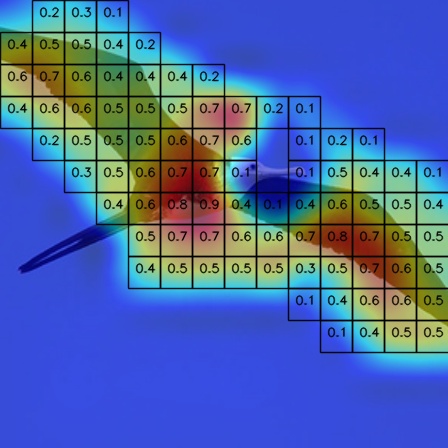}} & 
\raisebox{-0.5\height}{\includegraphics{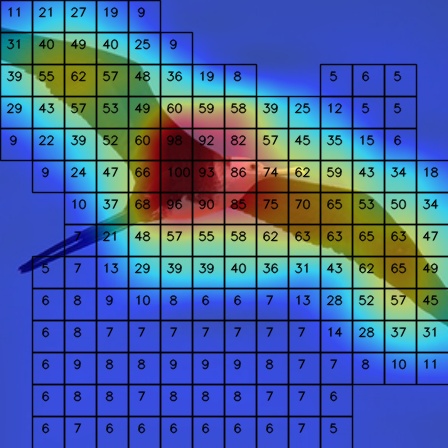}}\\
 & 99.3\%&99.3\%&99.3\%&99.3\%&41.3\%&41.3\%\\ 
 & Layer-CAM & FD-CAM & Lift-CAM & Score-CAM & CAPE (TS) & $\mu$-CAPE (TS)  \\
& 
\raisebox{-0.5\height}{\includegraphics{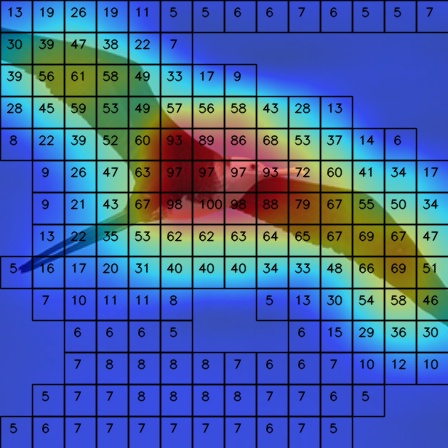}} &
\raisebox{-0.5\height}{\includegraphics{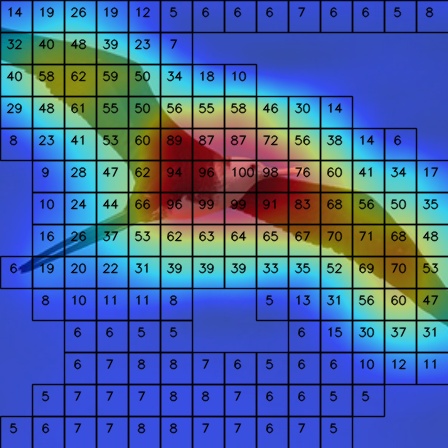}} &
\raisebox{-0.5\height}{\includegraphics{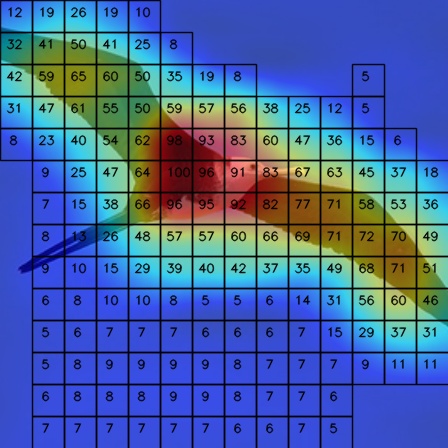}} & 
\raisebox{-0.5\height}{\includegraphics{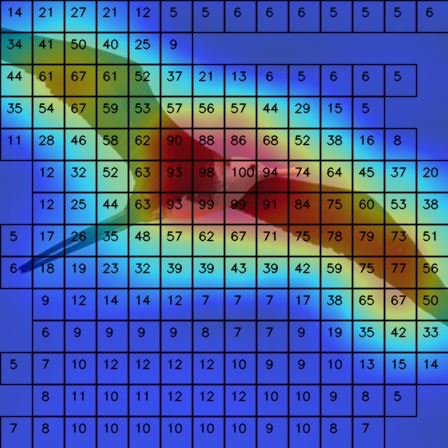}} & 
\raisebox{-0.5\height}{\includegraphics{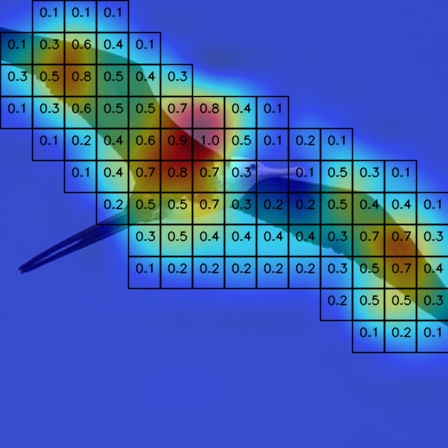}} & 
\raisebox{-0.5\height}{\includegraphics{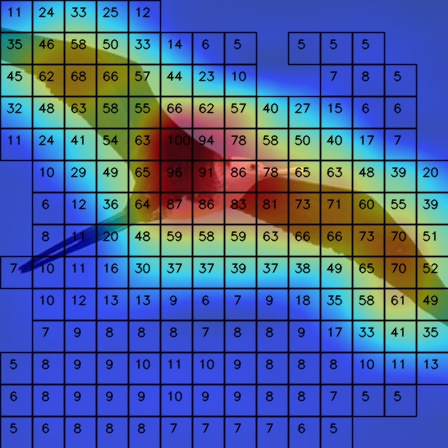}}\\
 & 99.3\%&99.3\%&99.3\%&99.3\%&31.1\%&31.1\%\\ 
 \\
& CAM & Grad-CAM & G-CAM++ & SG-CAM++ & CAPE (PF) & $\mu$-CAPE (PF) \\
\multirow{5}{*}{\raisebox{-0.5\height}{\shortstack{\includegraphics{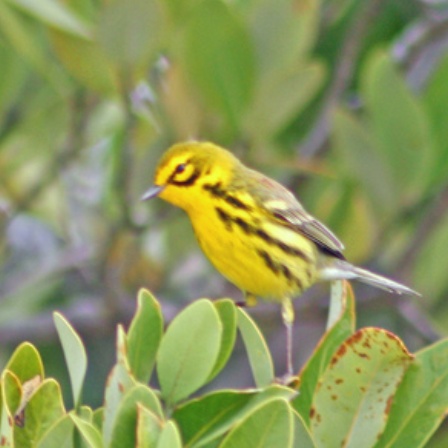}\\Palm Warbler}}} & 
\raisebox{-0.5\height}{\includegraphics{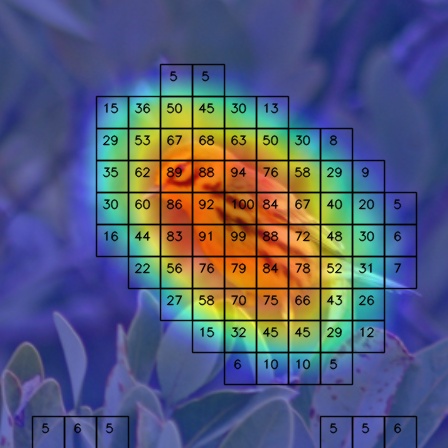}} & 
\raisebox{-0.5\height}{\includegraphics{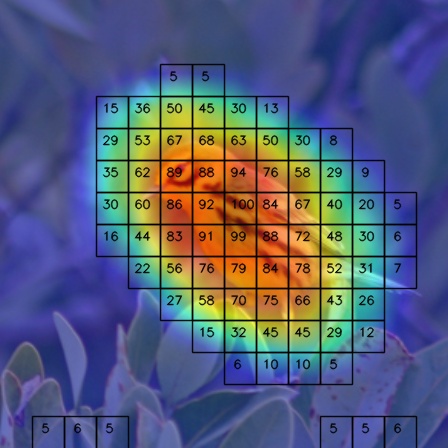}} & 
\raisebox{-0.5\height}{\includegraphics{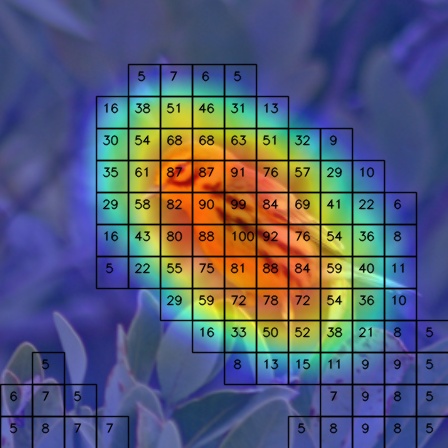}} & 
\raisebox{-0.5\height}{\includegraphics{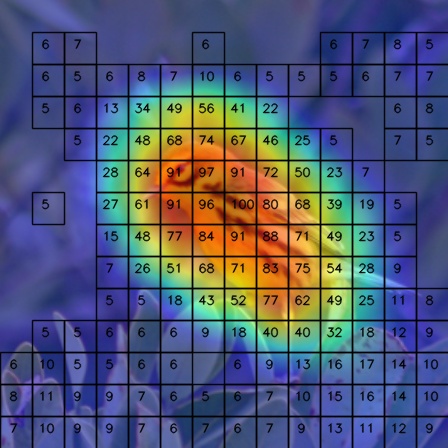}} & 
\raisebox{-0.5\height}{\includegraphics{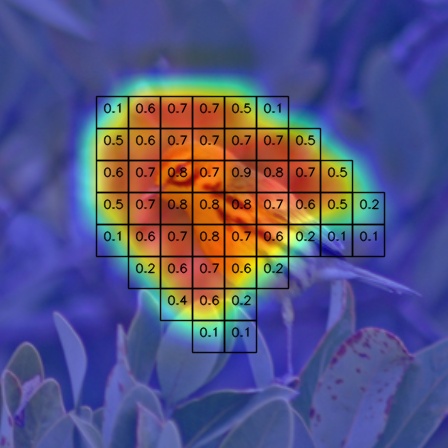}} & 
\raisebox{-0.5\height}{\includegraphics{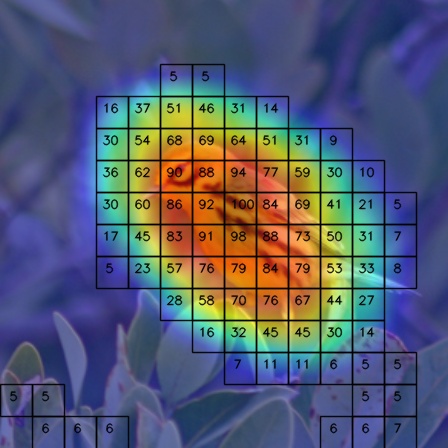}}\\
 & 92.6\%&92.6\%&92.6\%&92.6\%&26.7\%&26.7\%\\ 
 & Layer-CAM & FD-CAM & Lift-CAM & Score-CAM & CAPE (TS) & $\mu$-CAPE (TS)  \\
& 
\raisebox{-0.5\height}{\includegraphics{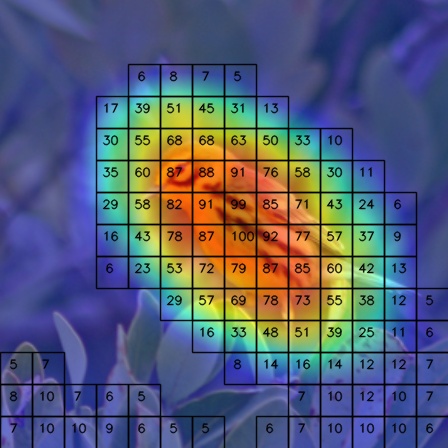}} &
\raisebox{-0.5\height}{\includegraphics{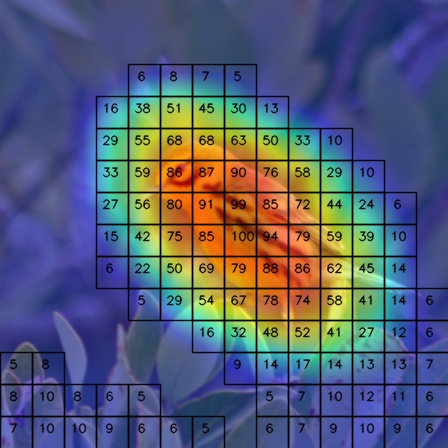}} &
\raisebox{-0.5\height}{\includegraphics{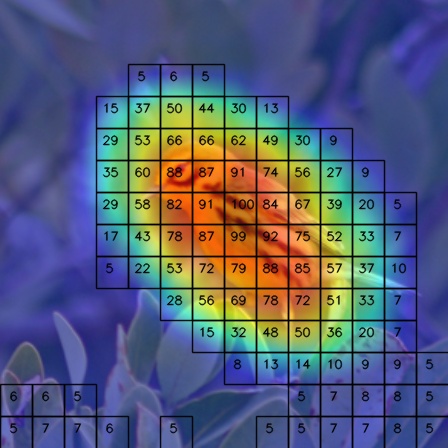}} & 
\raisebox{-0.5\height}{\includegraphics{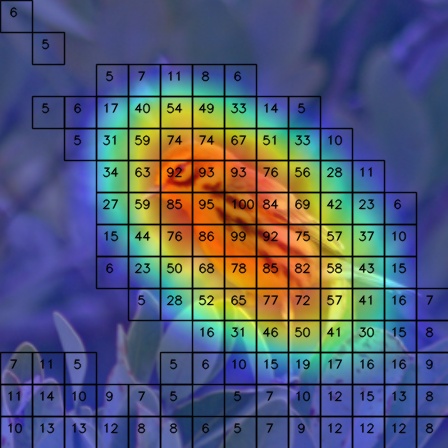}} & 
\raisebox{-0.5\height}{\includegraphics{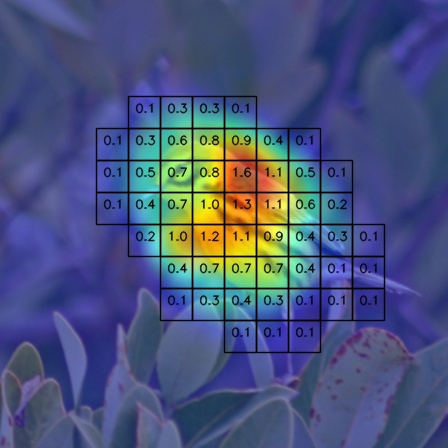}} & 
\raisebox{-0.5\height}{\includegraphics{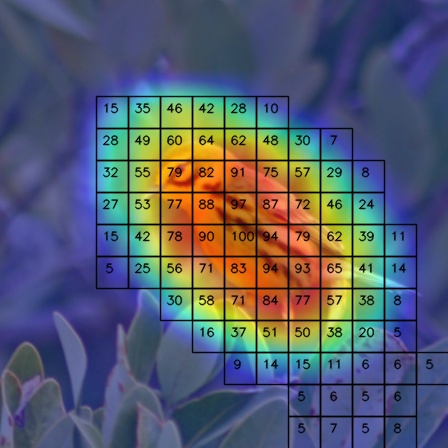}}\\
& 92.6\%&92.6\%&92.6\%&92.6\%&25.2\%&25.2\%\\ 
\\
 & CAM & Grad-CAM & G-CAM++ & SG-CAM++ & CAPE (PF) & $\mu$-CAPE (PF) \\
\multirow{5}{*}{\raisebox{-0.5\height}{\shortstack{\includegraphics{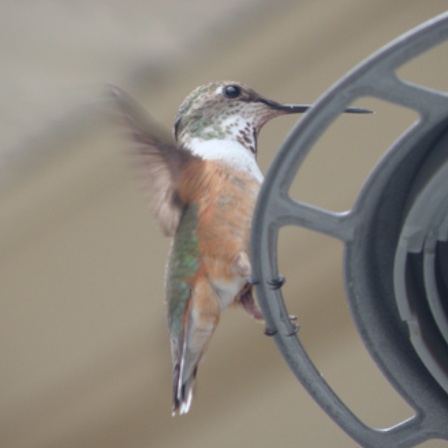}\\Anna\\Hummingbird}}} & 
\raisebox{-0.5\height}{\includegraphics{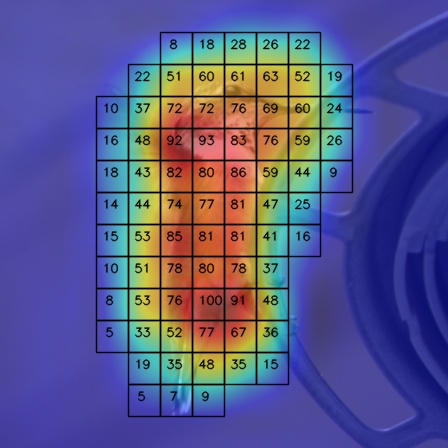}} & 
\raisebox{-0.5\height}{\includegraphics{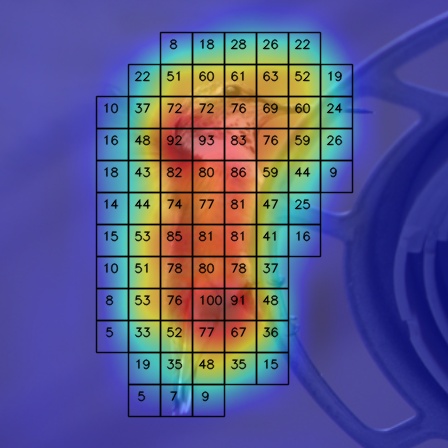}} & 
\raisebox{-0.5\height}{\includegraphics{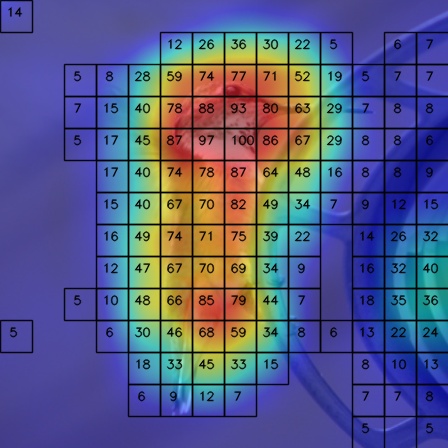}} & 
\raisebox{-0.5\height}{\includegraphics{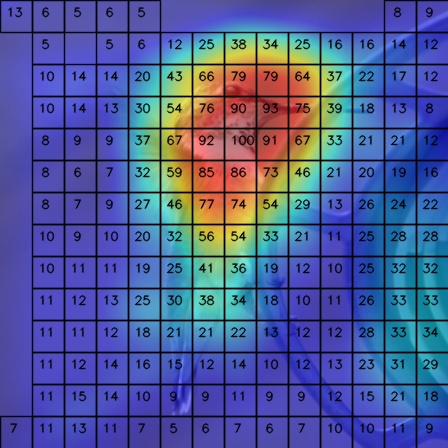}} & 
\raisebox{-0.5\height}{\includegraphics{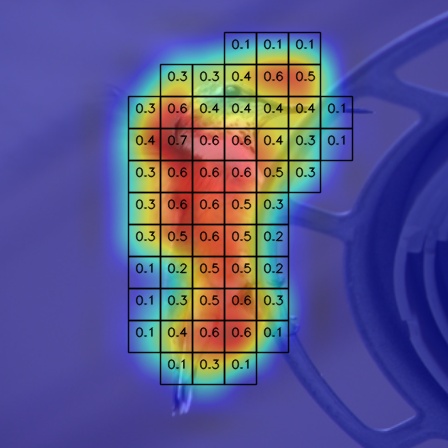}} & 
\raisebox{-0.5\height}{\includegraphics{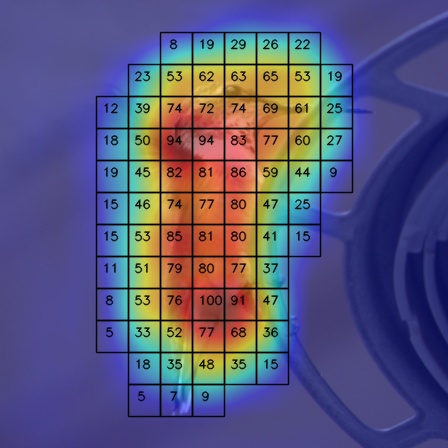}}\\
& 54.7\%&54.7\%&54.7\%&54.7\%&21.9\%&21.9\%\\ 
 & Layer-CAM & FD-CAM & Lift-CAM & Score-CAM & CAPE (TS) & $\mu$-CAPE (TS)  \\ 
 & 
\raisebox{-0.5\height}{\includegraphics{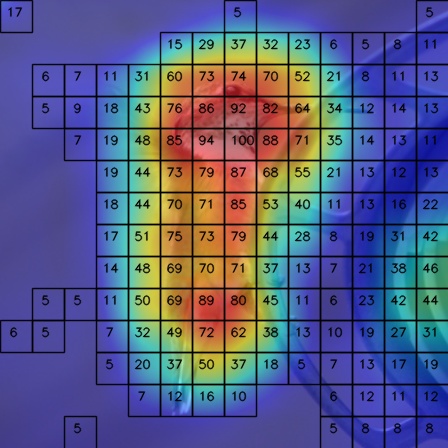}} &
\raisebox{-0.5\height}{\includegraphics{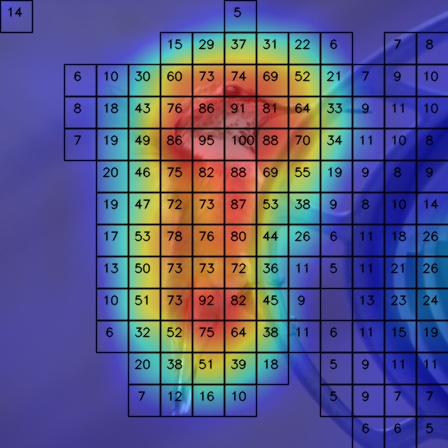}} &
\raisebox{-0.5\height}{\includegraphics{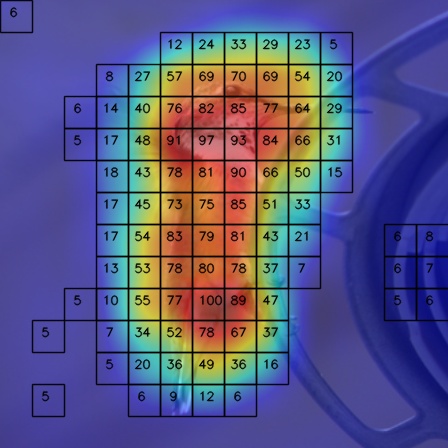}} & 
\raisebox{-0.5\height}{\includegraphics{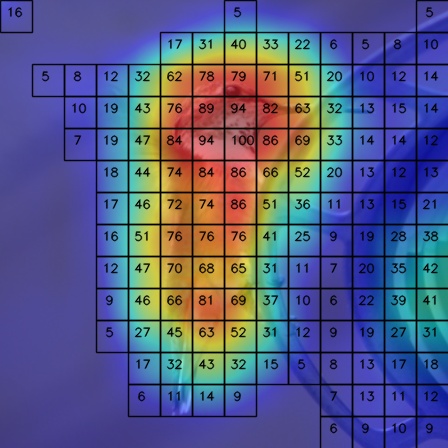}} & 
\raisebox{-0.5\height}{\includegraphics{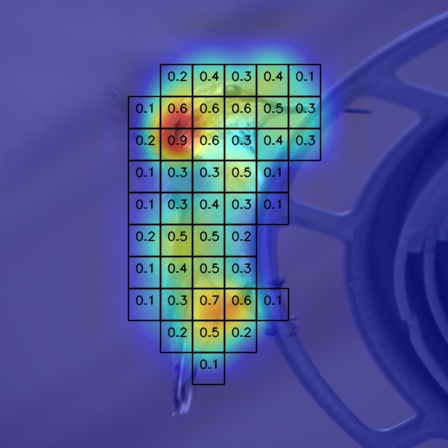}} & 
\raisebox{-0.5\height}{\includegraphics{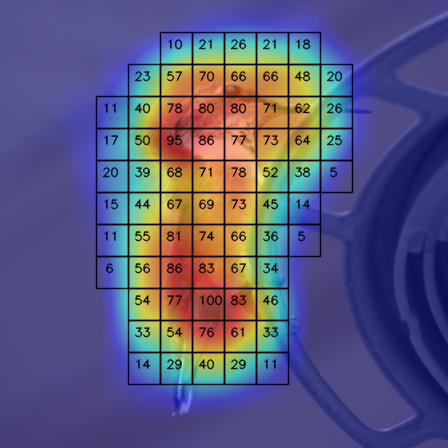}}\\
& 54.7\%&54.7\%&54.7\%&54.7\%&15.6\%&15.6\%\\ 

\end{tabularx}
}
\caption{Qualitative visualization using the ResNet-50 backbone model for CUB dataset. 
The class confidence scores are shown under the respective explanation maps.
``G-CAM++'' and ``SG-CAM++'' denote Grad-CAM++ and Smooth Grad-CAM++ respectively.}
\label{fig:qualitative_cub}
\end{figure*}

\begin{figure*}[!htbp]
\centering
\resizebox{1\textwidth}{!}{
\setkeys{Gin}{width=1.1\linewidth}
\newcolumntype{C}{>{\centering\arraybackslash}X}
\begin{tabularx}{\textwidth}{CCCCCCC}
Original & CAM & Grad-CAM & G-CAM++ & SG-CAM++ & CAPE (PF) & $\mu$-CAPE (PF) \\
\multirow{5}{*}{\raisebox{-0.5\height}{\shortstack{\includegraphics{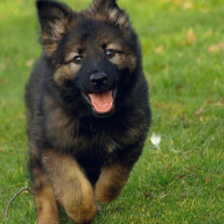}\\German\\Shepherd}}} & 
\raisebox{-0.5\height}{\includegraphics{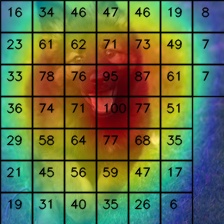}} & 
\raisebox{-0.5\height}{\includegraphics{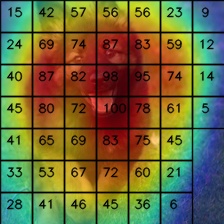}} & 
\raisebox{-0.5\height}{\includegraphics{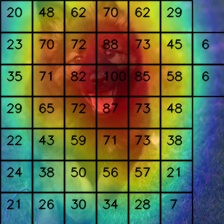}} & 
\raisebox{-0.5\height}{\includegraphics{figures/qualitative/imagenet/cam_0_SmoothGradCAMpp_cid_235_pred_73.93.jpg}} & 
\raisebox{-0.5\height}{\includegraphics{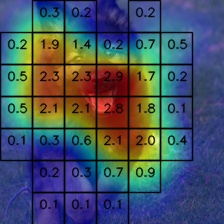}} & 
\raisebox{-0.5\height}{\includegraphics{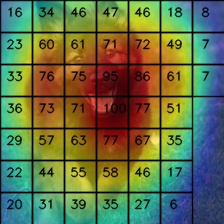}}\\
& 73.9\% & 73.9\% & 73.9\% & 73.9\% & 32.9\% & 32.9\%\\
& Layer-CAM & FD-CAM & Lift-CAM & Score-CAM & CAPE (TS) & $\mu$-CAPE (TS) \\
& 
\raisebox{-0.5\height}{\includegraphics{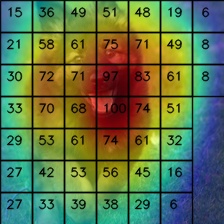}} &
\raisebox{-0.5\height}{\includegraphics{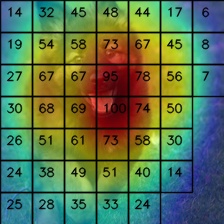}} &
\raisebox{-0.5\height}{\includegraphics{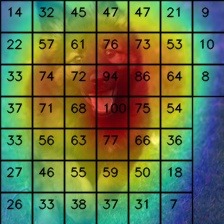}} & 
\raisebox{-0.5\height}{\includegraphics{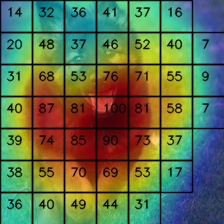}} & 
\raisebox{-0.5\height}{\includegraphics{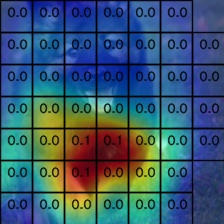}} & 
\raisebox{-0.5\height}{\includegraphics{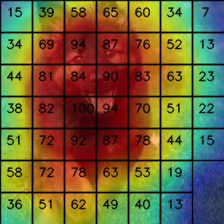}}\\
& 73.9\% & 73.9\% & 73.9\% & 73.9\% & 1.07\% & 1.07\%\\
\\
& CAM & Grad-CAM & G-CAM++ & SG-CAM++ & CAPE (PF) & $\mu$-CAPE (PF) \\
\multirow{5}{*}{\raisebox{-0.5\height}{\shortstack{\includegraphics{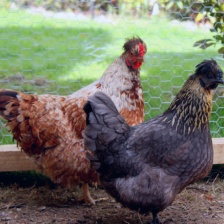}\\Hen}}} & 
\raisebox{-0.5\height}{\includegraphics{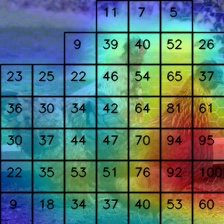}} & 
\raisebox{-0.5\height}{\includegraphics{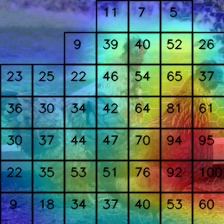}} & 
\raisebox{-0.5\height}{\includegraphics{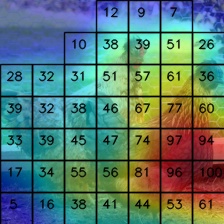}} & 
\raisebox{-0.5\height}{\includegraphics{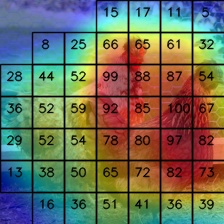}} & 
\raisebox{-0.5\height}{\includegraphics{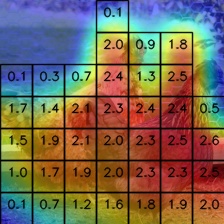}} & 
\raisebox{-0.5\height}{\includegraphics{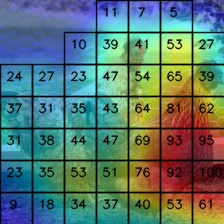}}\\
& 97.3\% & 97.3\% & 97.3\% & 97.3\% & 63.0\% & 63.0\%\\
& Layer-CAM & FD-CAM & Lift-CAM & Score-CAM & CAPE (TS) & $\mu$-CAPE (TS) \\
& 
\raisebox{-0.5\height}{\includegraphics{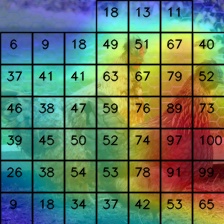}} &
\raisebox{-0.5\height}{\includegraphics{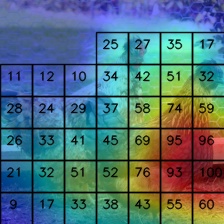}} &
\raisebox{-0.5\height}{\includegraphics{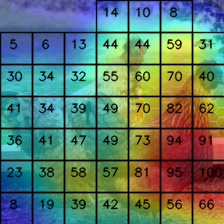}} & 
\raisebox{-0.5\height}{\includegraphics{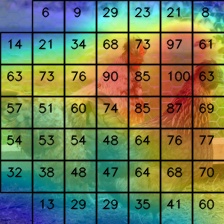}} & 
\raisebox{-0.5\height}{\includegraphics{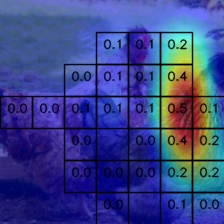}} & 
\raisebox{-0.5\height}{\includegraphics{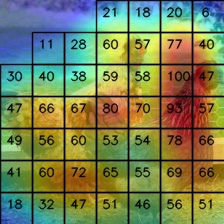}}\\
& 97.3\% & 97.3\% & 97.3\% & 97.3\% & 3.26\% & 3.26\%\\
\\

 & CAM & Grad-CAM & G-CAM++ & SG-CAM++ & CAPE (PF) & $\mu$-CAPE (PF) \\
\multirow{5}{*}{\raisebox{-0.5\height}{\shortstack{\includegraphics{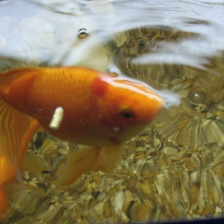}\\Goldfish}}} & 
\raisebox{-0.5\height}{\includegraphics{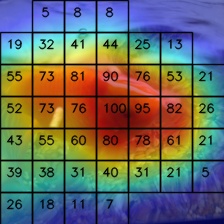}} & 
\raisebox{-0.5\height}{\includegraphics{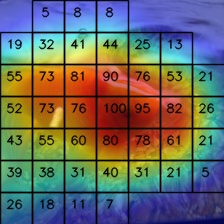}} & 
\raisebox{-0.5\height}{\includegraphics{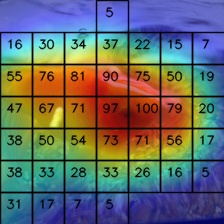}} & 
\raisebox{-0.5\height}{\includegraphics{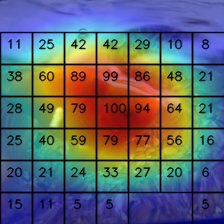}} & 
\raisebox{-0.5\height}{\includegraphics{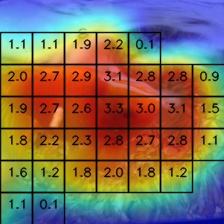}} &
\raisebox{-0.5\height}{\includegraphics{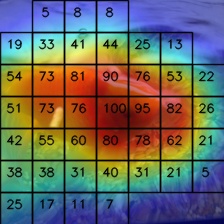}}
\\
& 99.9\% & 99.9\% & 99.9\% & 99.9\% & 68.1\% & 68.1\%\\
 & Layer-CAM & FD-CAM & Lift-CAM & Score-CAM & CAPE (TS) & $\mu$-CAPE (TS) \\
 & 
\raisebox{-0.5\height}{\includegraphics{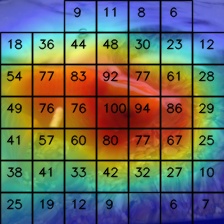}} &
\raisebox{-0.5\height}{\includegraphics{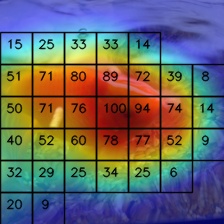}} &
\raisebox{-0.5\height}{\includegraphics{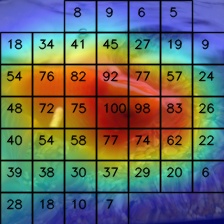}} & 
\raisebox{-0.5\height}{\includegraphics{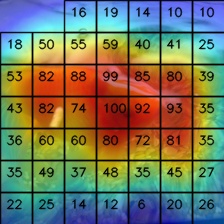}} & 
\raisebox{-0.5\height}{\includegraphics{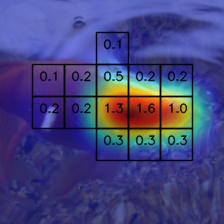}} &
\raisebox{-0.5\height}{\includegraphics{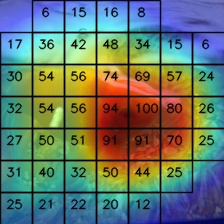}}\\
& 99.9\% & 99.9\% & 99.9\% & 99.9\% & 6.98\% & 6.98\%\\

\end{tabularx}
}
\caption{Qualitative visualization using the ResNet-50 backbone model for Imagenet. 
``G-CAM++'' and ``SG-CAM++'' denote Grad-CAM++ and Smooth Grad-CAM++ respectively.}
\label{fig:qualitative_imagenet}
\end{figure*}

\begin{figure*}[!htbp]
\centering
\resizebox{0.95\textwidth}{!}{
\setkeys{Gin}{width=1.1\linewidth}
\newcolumntype{C}{>{\centering\arraybackslash}X}
\begin{tabularx}{\textwidth}{CCCCCCC}%
Original & CAM & Grad-CAM & G-CAM++ & SG-CAM++ & CAPE (PF) & $\mu$-CAPE (PF) \\
&
\raisebox{-0.5\height}{\includegraphics{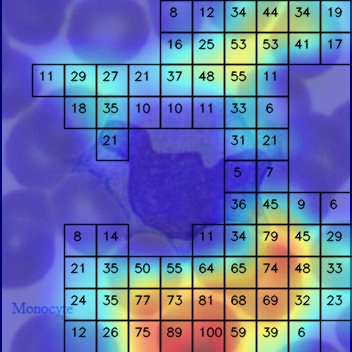}} & 
\raisebox{-0.5\height}{\includegraphics{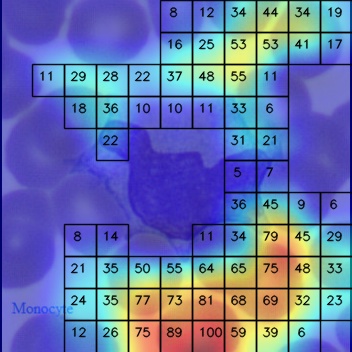}} & 
\raisebox{-0.5\height}{\includegraphics{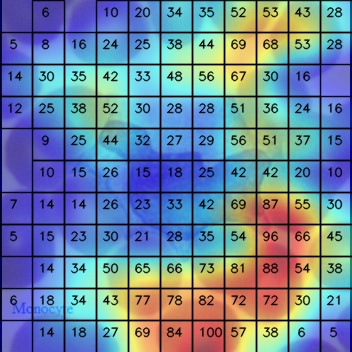}} & 
\raisebox{-0.5\height}{\includegraphics{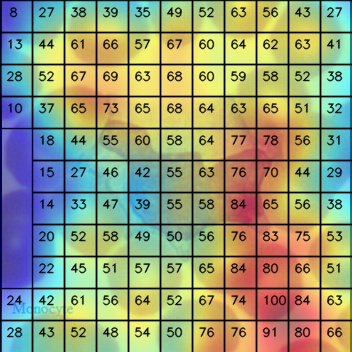}} & 
\raisebox{-0.5\height}{\includegraphics{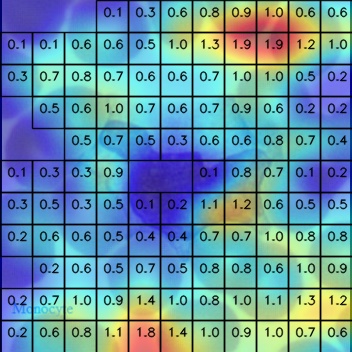}} &
\raisebox{-0.5\height}{\includegraphics{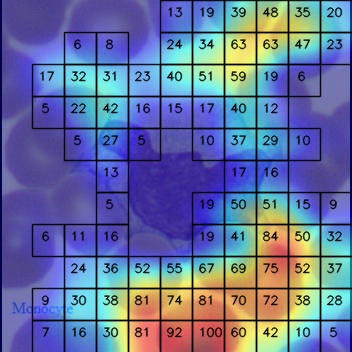}}\\
& Normal:99.9\% & Normal:99.9\% & Normal:99.9\% & Normal:99.9\% & Normal:76.6\% & Normal:76.6\% \\
& Layer-CAM & FD-CAM & Lift-CAM & Score-CAM & CAPE (TS) & $\mu$-CAPE (TS)  \\
\multirow{3}{*}{\raisebox{-0.5\height}{\shortstack{\includegraphics{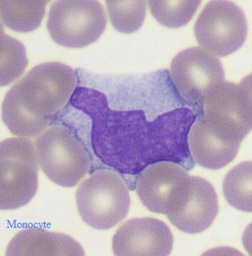}\\GT: Normal}}} & 
\raisebox{-0.5\height}{\includegraphics{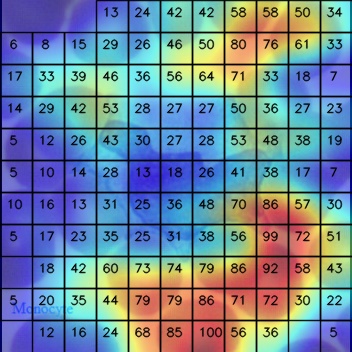}} & 
\raisebox{-0.5\height}{\includegraphics{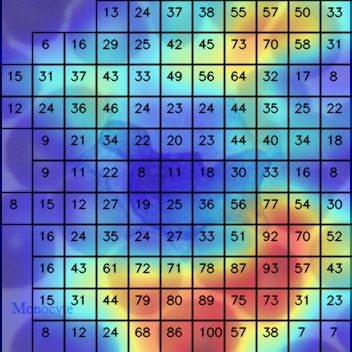}} & 
\raisebox{-0.5\height}{\includegraphics{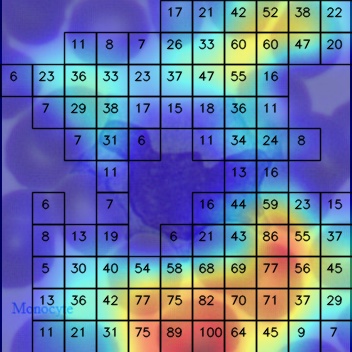}} & 
\raisebox{-0.5\height}{\includegraphics{figures/qualitative/cmml/0/cam_2_SmoothGradCAMpp_cid_0_pred_99.88.jpg}} & 
\raisebox{-0.5\height}{\includegraphics{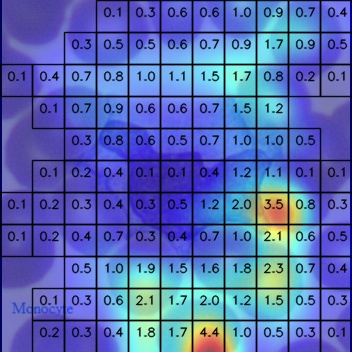}} &
\raisebox{-0.5\height}{\includegraphics{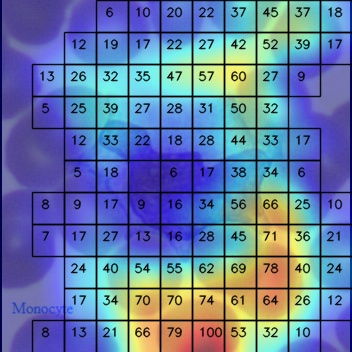}}\\
& Normal:99.9\% & Normal:99.9\% & Normal:99.9\% & Normal:99.9\% & Normal:84.6\% & Normal:84.6\% \\
 & & & & & & \\
& CAM & Grad-CAM & G-CAM++ & SG-CAM++ & CAPE (PF) & $\mu$-CAPE (PF) \\
& \raisebox{-0.5\height}{\includegraphics{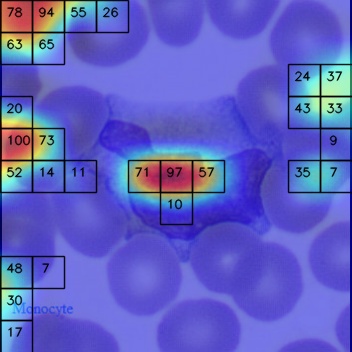}} & 
\raisebox{-0.5\height}{\includegraphics{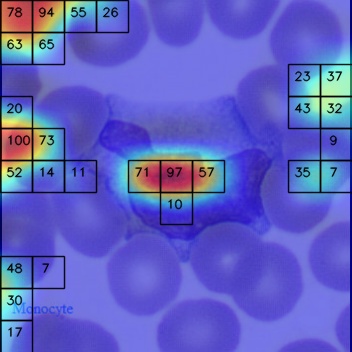}} & 
\raisebox{-0.5\height}{\includegraphics{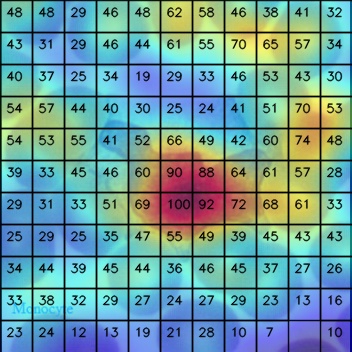}} & 
\raisebox{-0.5\height}{\includegraphics{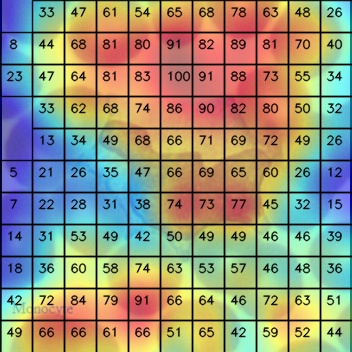}} & 
\raisebox{-0.5\height}{\includegraphics{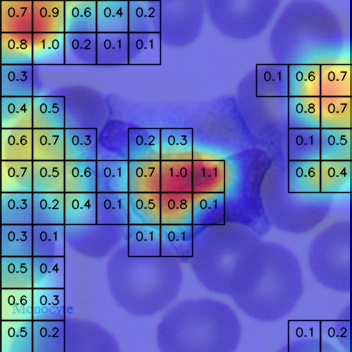}} &
\raisebox{-0.5\height}{\includegraphics{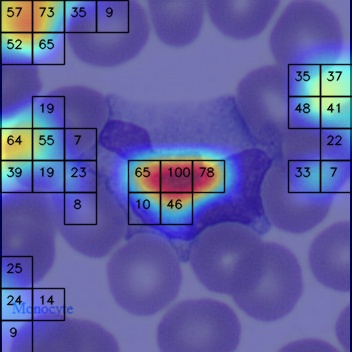}}\\
& CMML:0.1\% & CMML:0.1\% & CMML:0.1\% & CMML:0.1\% & CMML:23.4\% & CMML:23.4\% \\ 
& Layer-CAM & FD-CAM & Lift-CAM & Score-CAM & CAPE (TS) & $\mu$-CAPE (TS)  \\
& \raisebox{-0.5\height}{\includegraphics{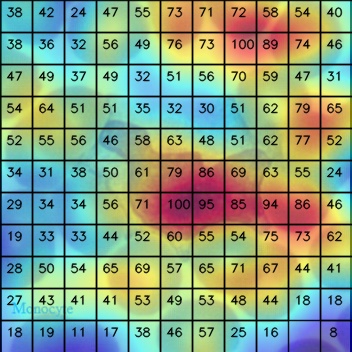}} & 
\raisebox{-0.5\height}{\includegraphics{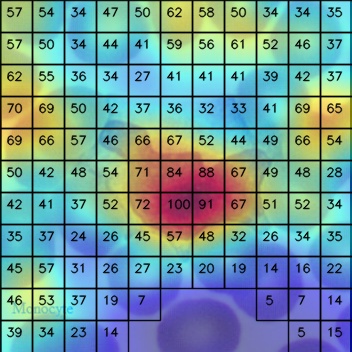}} & 
\raisebox{-0.5\height}{\includegraphics{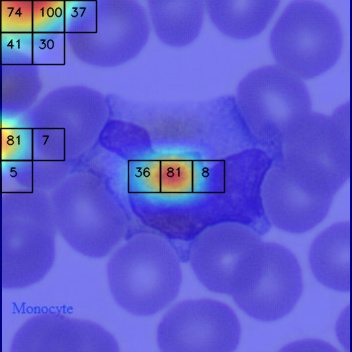}} & 
\raisebox{-0.5\height}{\includegraphics{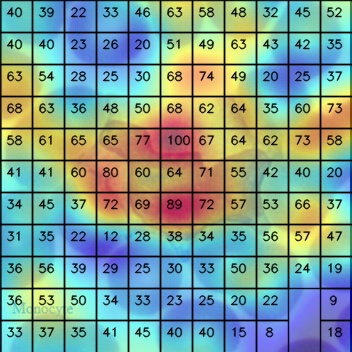}} & 
\raisebox{-0.5\height}{\includegraphics{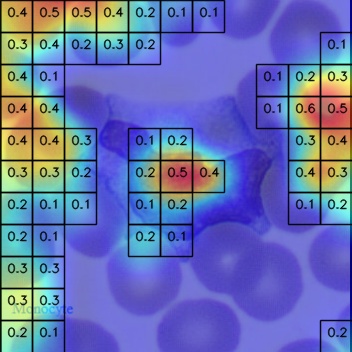}} &
\raisebox{-0.5\height}{\includegraphics{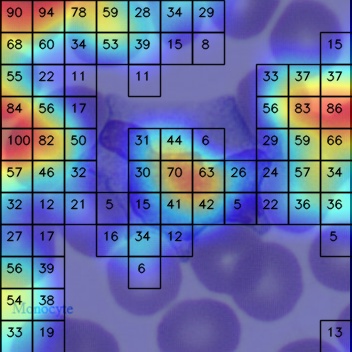}} \\
& CMML:0.1\% & CMML:0.1\% & CMML:0.1\% & CMML:0.1\% & CMML:15.4\% & CMML:15.4\% \\
\\
 & CAM & Grad-CAM & G-CAM++ & SG-CAM++ & CAPE (PF) & $\mu$-CAPE (PF) \\
&
\raisebox{-0.5\height}{\includegraphics{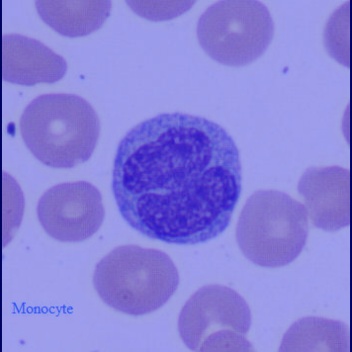}} & 
\raisebox{-0.5\height}{\includegraphics{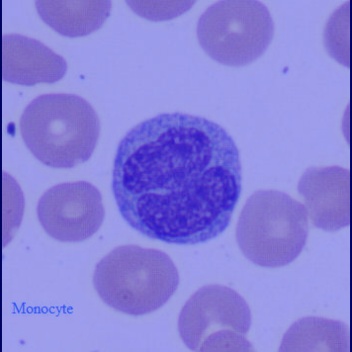}} & 
\raisebox{-0.5\height}{\includegraphics{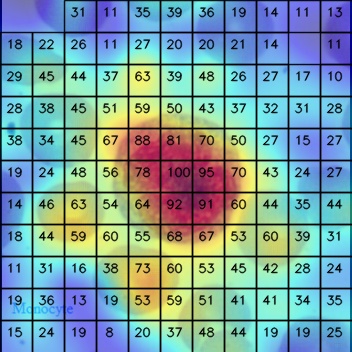}} & 
\raisebox{-0.5\height}{\includegraphics{figures/qualitative/cmml/TS/0/cam_1_GradCAMpp_cid_1_pred_0.0.jpg}} & 
\raisebox{-0.5\height}{\includegraphics{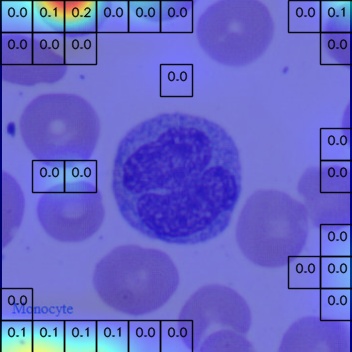}} &
\raisebox{-0.5\height}{\includegraphics{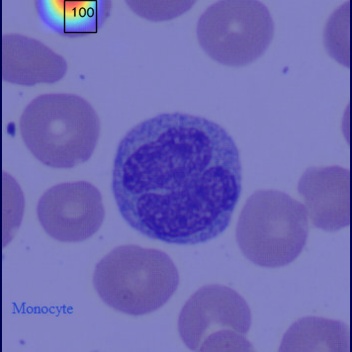}}\\
& Normal:0.0\% & Normal:0.0\% & Normal:0.0\% & Normal:0.0\% & Normal:1.2\% & Normal:1.2\% \\
& Layer-CAM & FD-CAM & Lift-CAM & Score-CAM & CAPE (TS) & $\mu$-CAPE (TS)  \\
\multirow{3}{*}{\raisebox{-0.5\height}{\shortstack{\includegraphics{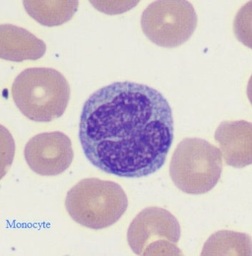}\\GT: CMML}}} & 
\raisebox{-0.5\height}{\includegraphics{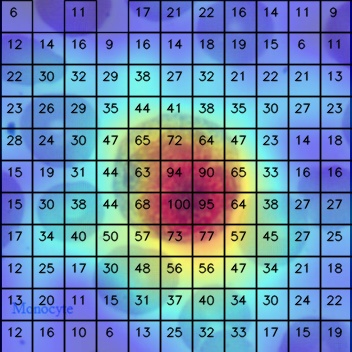}} & 
\raisebox{-0.5\height}{\includegraphics{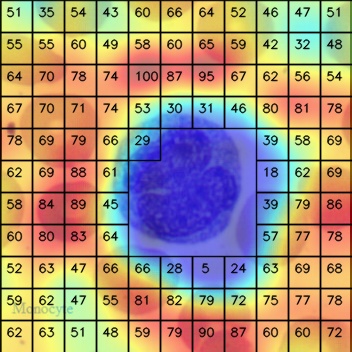}} & 
\raisebox{-0.5\height}{\includegraphics{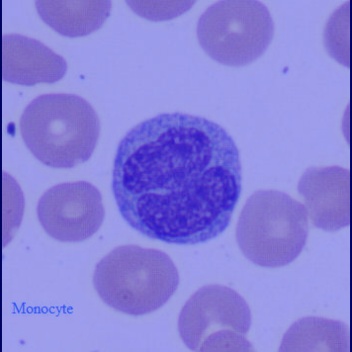}} & 
\raisebox{-0.5\height}{\includegraphics{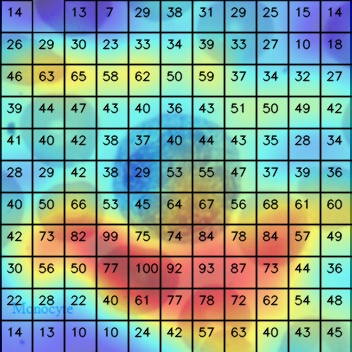}} & 
\raisebox{-0.5\height}{\includegraphics{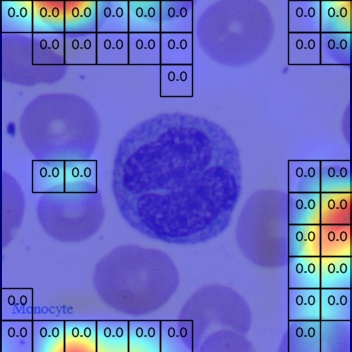}} &
\raisebox{-0.5\height}{\includegraphics{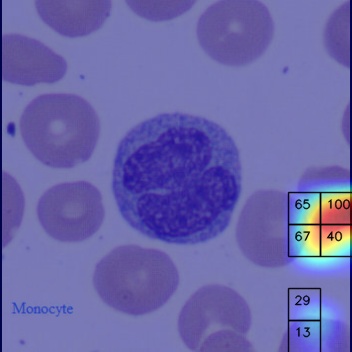}}\\
& Normal:0.0\% & Normal:0.0\% & Normal:0.0\% & Normal:0.0\% & Normal:0.2\% & Normal:0.2\% \\
 & & & & & & \\
& CAM & Grad-CAM & G-CAM++ & SG-CAM++ & CAPE (PF) & $\mu$-CAPE (PF) \\
& \raisebox{-0.5\height}{\includegraphics{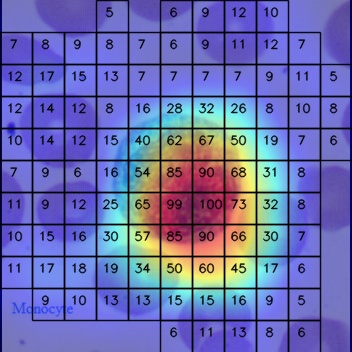}} & 
\raisebox{-0.5\height}{\includegraphics{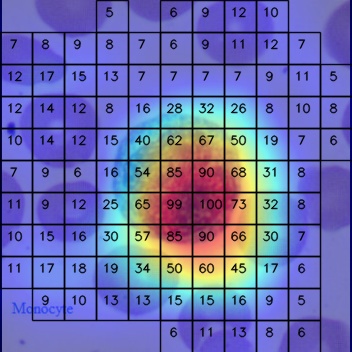}} & 
\raisebox{-0.5\height}{\includegraphics{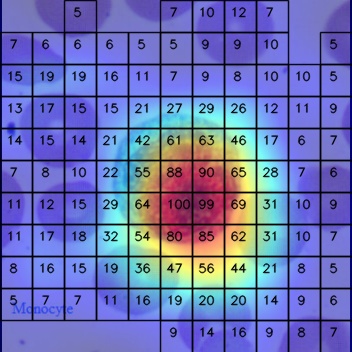}} & 
\raisebox{-0.5\height}{\includegraphics{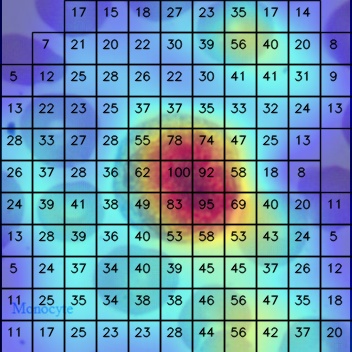}} & 
\raisebox{-0.5\height}{\includegraphics{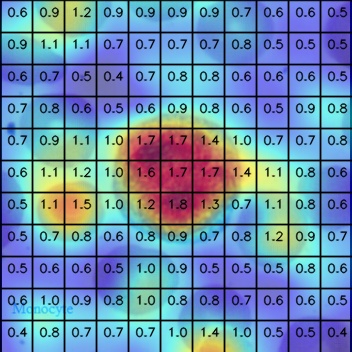}} &
\raisebox{-0.5\height}{\includegraphics{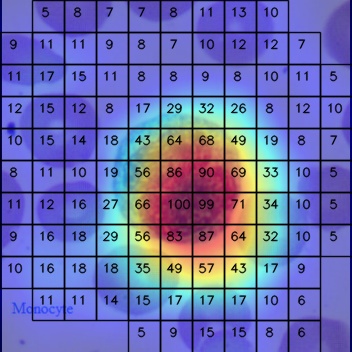}}\\
& CMML:100.0\% & CMML:100.0\% & CMML:100.0\% & CMML:100.0\% & CMML:98.8\% & CMML:98.8\% \\ 
& Layer-CAM & FD-CAM & Lift-CAM & Score-CAM & CAPE (TS) & $\mu$-CAPE (TS)  \\
& \raisebox{-0.5\height}{\includegraphics{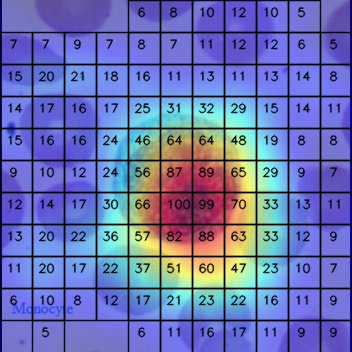}} & 
\raisebox{-0.5\height}{\includegraphics{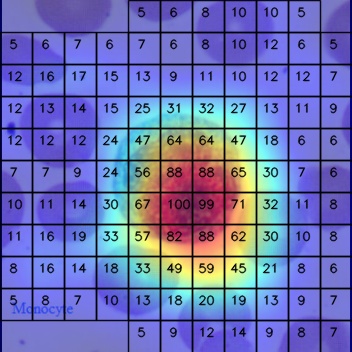}} & 
\raisebox{-0.5\height}{\includegraphics{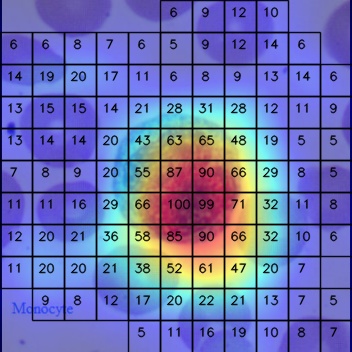}} & 
\raisebox{-0.5\height}{\includegraphics{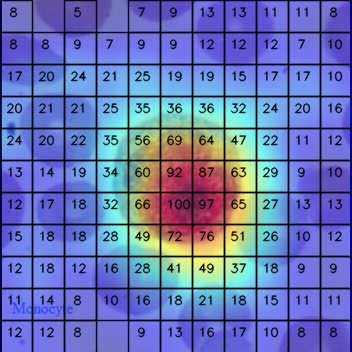}} & 
\raisebox{-0.5\height}{\includegraphics{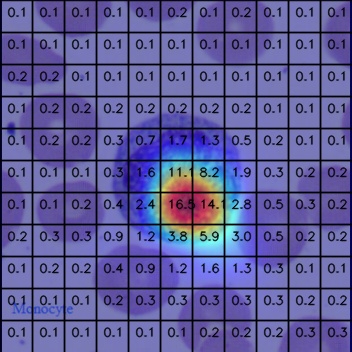}} &
\raisebox{-0.5\height}{\includegraphics{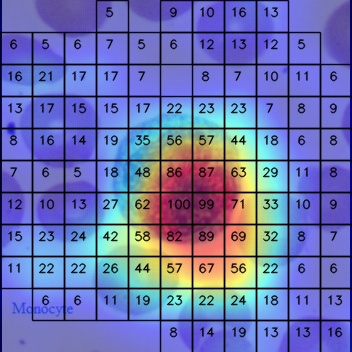}} \\
& CMML:100.0\% & CMML:100.0\% & CMML:100.0\% & CMML:100.0\% & CMML:99.9\% & CMML:99.9\% \\

\end{tabularx}
}
\caption{Qualitative visualization using the ResNet-50 backbone for one Normal example (top) and one CMML example (bottom).} 
\label{fig:qualitative_cmml}
\end{figure*}

Due to the limited space in the main manuscript, we show the full qualitative examples and comparisons across all eight state-of-the-art CAM maps in this supplementary material.
The additional samples are shown in Fig.~\ref{fig:qualitative_cub}, Fig.~\ref{fig:qualitative_imagenet}, and Fig.~\ref{fig:qualitative_cmml} for CUB, ImageNet, and CMML respectively, at the end of this document.
The comparison between CAPE (PF) and (TS) suggests that (PF) CAPE generally gives a larger region of attention and accumulatively less softened class prediction, which is aligned with the observation in Table~\ref{tab:prediction_confidence} (see the rows for Bootstrap (PF) and (TS)).
In Fig.~\ref{fig:qualitative_cmml}, the CAM, Grad-CAM, and Lift-CAM do not yield any attention for the CMML example.
This is because these methods have produced all negative values for the respective CAM and the rectifier function clipped the values to zero, hence not showing any attention.

\begin{table}[!ht]
\centering
\resizebox{\columnwidth}{!}{
\begin{tabular}{c|ccccc|c}%
\hline
\multirow{2}{*}{Method} & \multicolumn{6}{c}{CUB}\\
\cline{2-7}
 & AD $\downarrow$ & IC $\uparrow$ & ADD $\uparrow$ & ADCC $\uparrow$ & mIoU $\downarrow$ & BC $\uparrow$\\
\hline
CAM~\cite{zhou2016learning} & \cellcolor{LimeGreen}3.5 & \cellcolor{SpringGreen}49.7 & 27.7 & 54.9 & 96.79 & 3 \\
Grad-CAM~\cite{selvaraju2017grad} & \cellcolor{ForestGreen}3.4 & \cellcolor{LimeGreen}50.1 & 29.2 & 56.3 & 96.79 & \cellcolor{SpringGreen}4 \\
Grad-CAM++~\cite{chattopadhay2018grad} & 4.2 & 47.5 & 26.3 & 58.4 & 95.69 & 0 \\
Layer-CAM~\cite{jiang2021layercam} & \cellcolor{LimeGreen}3.5 & 48.4 & 28.6 & 56.4 & 97.06 & 2 \\
Score-CAM~\cite{wang2020score} & 6.5 & 46.1 & \cellcolor{ForestGreen}46.5 & \cellcolor{ForestGreen}78.1 & \cellcolor{SpringGreen}43.95 &  \cellcolor{ForestGreen}7 \\
\hline
CAPE (PF) & 14.3 & 33.7 & 22.7 & \cellcolor{SpringGreen}71.4 & \cellcolor{ForestGreen}17.19 &  \cellcolor{SpringGreen}4 \\
CAPE (TS) & 21.9 & 22.6 & 19.6 & \cellcolor{LimeGreen}73.4 &  \cellcolor{LimeGreen}21.34 & \cellcolor{SpringGreen}4 \\
\hline
$\mu$-CAPE (PF) & \cellcolor{SpringGreen}4.1 & \cellcolor{ForestGreen}52.7 & \cellcolor{LimeGreen}40.3 & 55.0 & 94.25 & \cellcolor{LimeGreen}6 \\
$\mu$-CAPE (TS) & 4.2 & 47.6 & \cellcolor{SpringGreen}37.6 & 54.0 & 96.96 & 1 \\
\hline

\end{tabular}
}
\caption{Comparison of different CAM interpretation methods for CUB using Swin Transformer V2-B as the DNN architecture. $\downarrow$ and $\uparrow$ indicate lower or higher is better. The top-3 scores are marked from darker to lighter green colors.}
\label{tab:quantitative_swin}
\end{table}

\begin{table}[!tbp]
\centering
\begin{tabular}{c|c}
\hline
Method & CUB \\
\hline
Naive AVG & 86.75 \\
Off-the-shelf & 87.15 \\
Bootstrap (TS) & 86.83 \\
Bootstrap (PF) & 87.14 \\
\hline
Vanilla Classifier & 87.12 \\
\hline
\end{tabular}
\caption{Accuracy comparison for Swin Transformer V2-B model on CUB.}
\label{tab:acc_swin}
\end{table}

\begin{table*}[!htbp]
    \centering
    \resizebox{\textwidth}{!}{
    \begin{tabular}{cc}
        \begin{tabular}{|c|c|c|c|}
        \hline
             & Normal & CMML & Total \\
        \hline
        Training set & 57 (928) & 14 (616) & 71 (1544) \\
        \hline
        Validation set & 40 (748) & 10 (472) & 40 (1220) \\
        \hline
        Test set & 40 (648) & 10 (487) & 50 (1,135) \\
        \hline
        Total & 137 (2,324) & 34 (1,575) & 171 (3,899) \\
        \hline
        \end{tabular}
         &  
        \begin{tabular}{|c|c|c|c|c|}
        \hline
	   &	BM Diagnosis	&	Observer 1	&	Observer 2	&	Observer 3	\\ \hline
        BM Diagnosis	&	100.0	&	59.5	&	49.7	&	48.4	\\ \hline
        Observer 1	&	59.5	&	100.0	&	52.3	&	65.4	\\ \hline
        Observer 2	&	49.7	&	52.3	&	100.0	&	55.6	\\ \hline
        Observer 3	&	48.4	&	65.4	&	55.6	&	100.0	\\ \hline
        \end{tabular} \\
        (a) Data statistics &
        (b) Human performance \& variability \\
    \end{tabular}
        
    }
    \caption{(a) Data statistics and (b) Human observer accuracy (\%) against the bone marrow (BM) diagnosis and inter-observer agreement.}
    \label{fig:dataset}
\end{table*}

\begin{table*}[!htbp]
\centering
\resizebox{0.9\textwidth}{!}{
\begin{tabular}{c|c|c|c|c|c|c|c}
\hline 
\multicolumn{2}{c|}{Semantic} & \multirow{2}{*}{Nucleus} &	Nuc/Cyto & \multirow{2}{*}{Cytoplasm} &	Cyto-Ext & Cell  &	Nuc/Cyto/Ext	\\
\multicolumn{2}{c|}{Class $\rightarrow$} &	 &	Boundary &	& Boundary	& Exterior	&	Boundary \\ \hline
\multicolumn{2}{c|}{Simplex Definition $\rightarrow$} & \multirow{2}{*}{$(100, 0, 0)$} & \multirow{2}{*}{$(50, 50, 0)$} & \multirow{2}{*}{$(0, 100, 0)$} & \multirow{2}{*}{$(0, 0, 100)$} & \multirow{2}{*}{$(0, 50, 50)$} & \multirow{2}{*}{$(33, 33, 33)$} \\ \cline{1-2}
Method & Class &  & &&&&\\ \hline
\multirow{3}{*}{CAPE (TS)} & Normal & 6.5$\pm$8.8	&	4.6$\pm$6.0	&	5.5$\pm$7.0	&	5.1$\pm$5.2	&	36.8$\pm$23.8	&	1.4$\pm$2.4\\ 
\cline{2-8}
&	CMML & 11.8$\pm$16.3	&	4.0$\pm$5.5	&	2.6$\pm$3.9	&	2.1$\pm$2.9	&	18.7$\pm$17.4	&	0.9$\pm$1.8\\ 
\cline{2-8}
&	CMML-Normal	& 5.3$\pm$22.0	&	-0.6$\pm$10.0	&	-2.9$\pm$9.3	&	-2.9$\pm$7.0	&	-18.1$\pm$37.1	&	-0.4$\pm$3.4\\
\hline
\multirow{3}{*}{CAPE (PF)} & Normal & 6.5$\pm$7.8	&	4.0$\pm$4.7	&	4.2$\pm$4.3	&	3.6$\pm$3.1	&	38.4$\pm$19.1	&	1.0$\pm$1.6\\ 
\cline{2-8}
&	CMML & 5.9$\pm$5.9	&	3.2$\pm$3.5	&	2.8$\pm$3.5	&	2.5$\pm$3.0	&	27.0$\pm$19.1	&	0.9$\pm$1.4\\
\cline{2-8}
&	CMML-Normal	& -0.6$\pm$12.7	&	-0.8$\pm$7.4	&	-1.4$\pm$6.9	&	-1.1$\pm$5.4	&	-11.4$\pm$36.9	&	-0.1$\pm$2.4\\
\hline

\end{tabular}
}
\caption{The image region contributions ($\%$) to 6 pre-defined semantic classes using the CAPE ResNet-50 model on all test images. For each method, the 12 (mean) contributions from the Normal/CMML class and the semantic class combinations sum to 100\%. }
\label{tab:semantic_contribution}
\end{table*}

\section{Experiments on Swin Transformer Model}

We trained the Swin V2-B transformer on CUB.
The training configuration for the CUB dataset is the same as the ResNet-50 model for the CUB dataset, except the batch size is set to 16 to cope with the larger GPU memory usage of the Swin Transformer model. 
We report the accuracy comparison of the Swin Transformer~\cite{liu2022swin}, vanilla classification model, and CAPE methods in Table~\ref{tab:acc_swin}.

\subsection{Quantitative Analysis}

We compare our method with five state-of-the-art CAM methods (CAM~\cite{zhou2016learning}, Layer-CAM~\cite{jiang2021layercam}, Score-CAM~\cite{wang2020score}, Grad-CAM~\cite{selvaraju2017grad}, Grad-CAM++~\cite{chattopadhay2018grad}) on CUB and ImageNet dataset. 
Table \ref{tab:quantitative_swin} presents the quantitative analysis using the same evaluation metrics in the main manuscript.  

On CUB, the classification performance of the CAPE models is very close to the vanilla classifier even without training, shown in Table~\ref{tab:acc_swin}.
In particular, the Off-the-shelf CAPE and Bootstrap (PF) models ($87.15\%$ and $87.12\%$) marginally surpass the performance of the vanilla classifier ($87.12\%$). 
In contrast, the performance gap between the Vanilla Classifier and Bootstrap (PF) on ResNet-50 was 1.22\%, the Vanilla Classifier was better.
Finally, the Bootstrap (TS) resulted in a lower performance of $86.83\%$, suggesting the full training course is unnecessary.

\subsection{Qualitative Analysis}

\begin{figure*}[!htbp]
\centering
\setkeys{Gin}{width=1.12\linewidth}
\newcolumntype{C}{>{\centering\arraybackslash}X}
\begin{tabularx}{\textwidth}{cCCCCCC}%
& Original & CAM & Grad-CAM++ & Score-CAM & CAPE (PF) & $\mu$-CAPE (PF) \\
 & 
\raisebox{-0.5\height}{\includegraphics{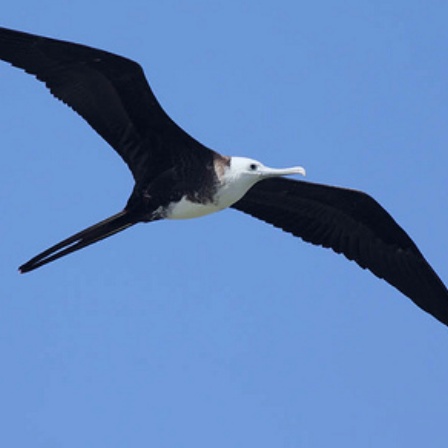}} & 
\raisebox{-0.5\height}{\includegraphics{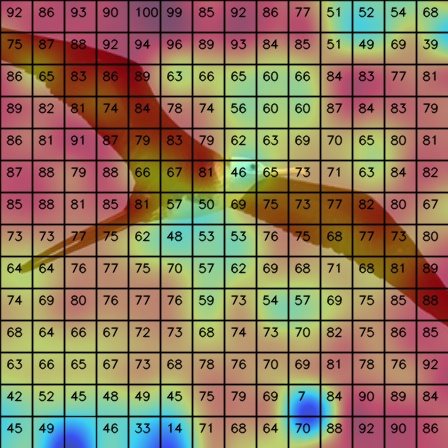}} & 
\raisebox{-0.5\height}{\includegraphics{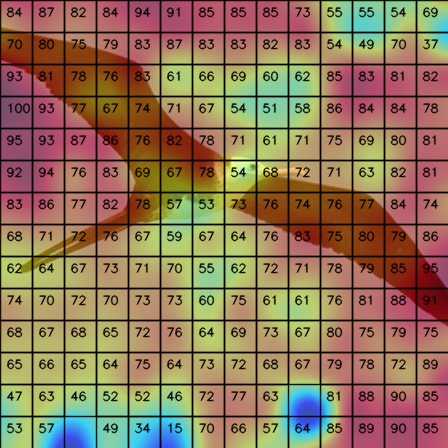}} & 
\raisebox{-0.5\height}{\includegraphics{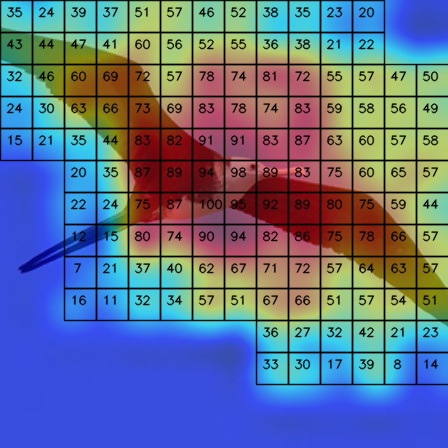}} & 
\raisebox{-0.5\height}{\includegraphics{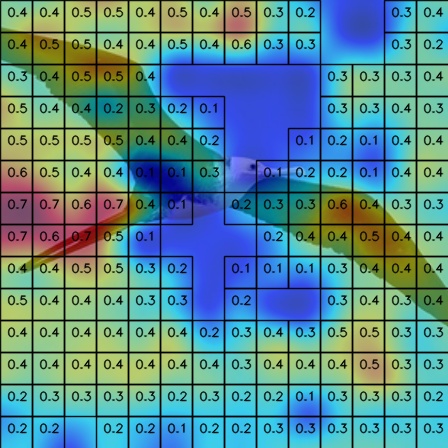}} &
\raisebox{-0.5\height}{\includegraphics{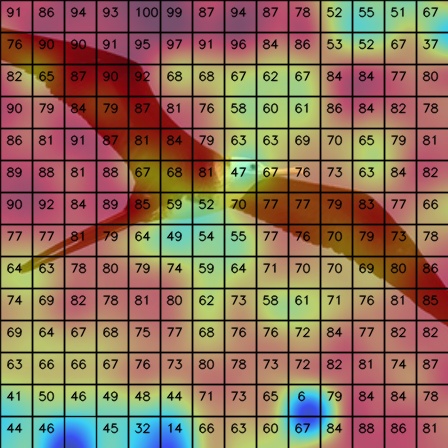}}
\\
& Frigatebird &95.9\%&95.9\%&95.9\%&60.2\%&60.2\%\\
\multirow{4}{*}{\rotatebox[origin=c]{90}{CUB }} & 
\raisebox{-0.5\height}{\includegraphics{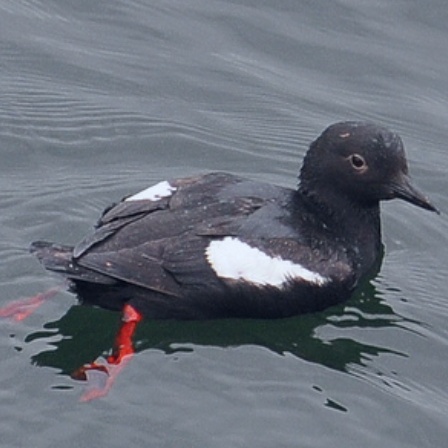}} & 
\raisebox{-0.5\height}{\includegraphics{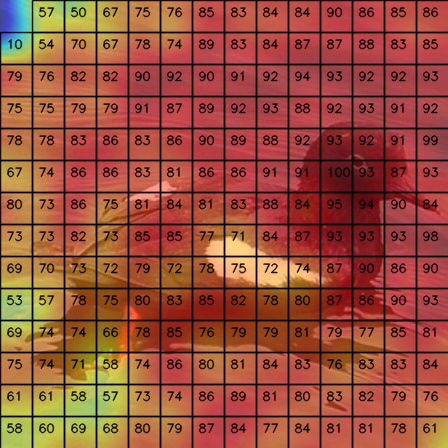}} & 
\raisebox{-0.5\height}{\includegraphics{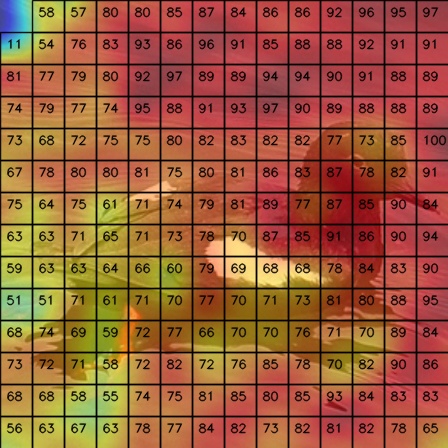}} & 
\raisebox{-0.5\height}{\includegraphics{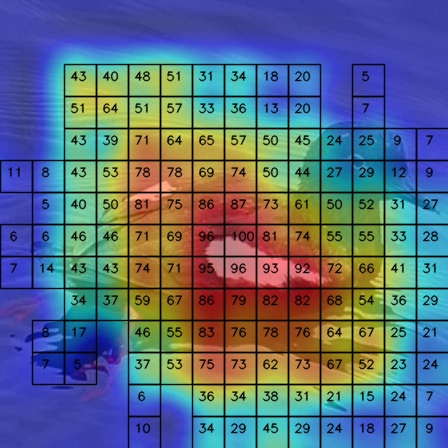}} & 
\raisebox{-0.5\height}{\includegraphics{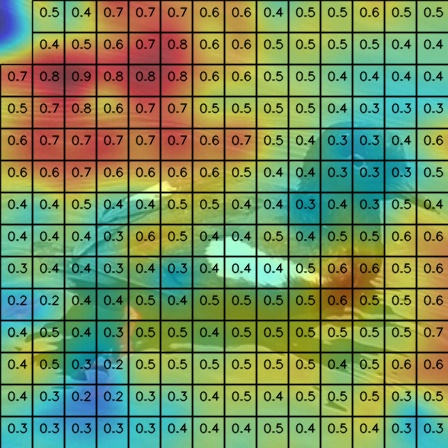}} &
\raisebox{-0.5\height}{\includegraphics{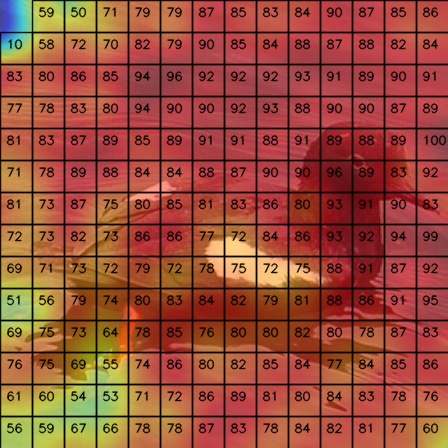}}
\\
& Pigeon Guillemot &100.0\%&100.0\%&100.0\%&93.5\%&93.5\%\\
& \raisebox{-0.5\height}{\includegraphics{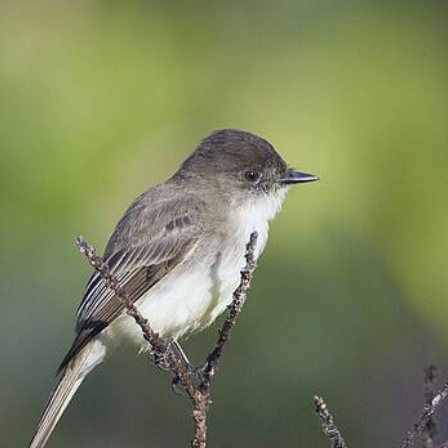}} & 
\raisebox{-0.5\height}{\includegraphics{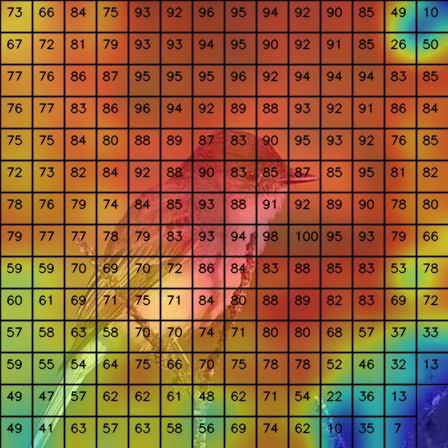}} & 
\raisebox{-0.5\height}{\includegraphics{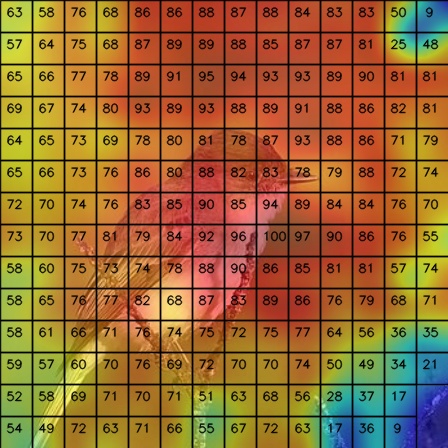}} & 
\raisebox{-0.5\height}{\includegraphics{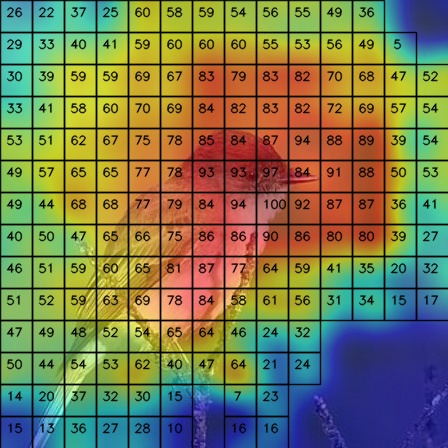}} & 
\raisebox{-0.5\height}{\includegraphics{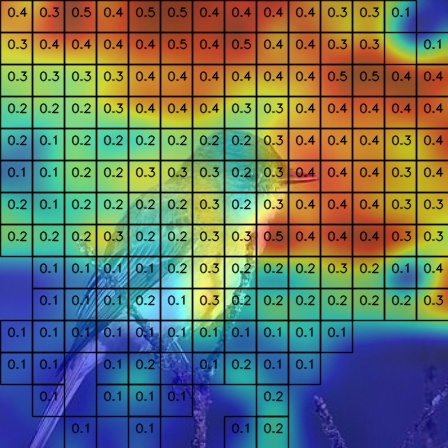}} &
\raisebox{-0.5\height}{\includegraphics{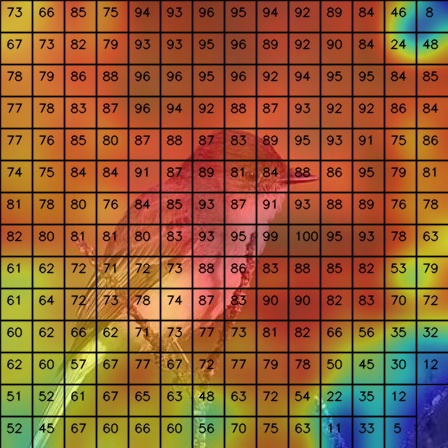}}
\\
& Sayornis &85.1\%&85.1\%&85.1\%&43.1\%&43.1\%\\

\end{tabularx}
\caption{Qualitative analysis for the CUB dataset using the Swin V2-B model. The class confidence scores are shown under the respective explanation maps, where CAM, Grad-CAM++, and Score-CAM visualize for the original classification model. CAPE and $\mu$-CAPE visualize for the post-fitted (PF) CAPE classification layer. Note that the shown values pre-upsampling values where we omit values $<0.5\%$ for CAPE and $\mu$-CAPE and $<5\%$ for the other CAMs.}
\label{fig:qualitative_swin}
\end{figure*}

Fig. \ref{fig:qualitative_swin} shows the visualization of different CAM methods for the Swin V2-B. 
It is clear that the model attention examples in Fig. \ref{fig:qualitative_swin} are all widespread. 
We suspect that this is due to the fact that the Transformer model tokenizes the image into non-overlapping patches and processes at the patch level, the spatial correspondence between the original input and the output CAM becomes weak.
This means that all patches in a transformer layer can access information of all patches in the layer below.
With the large model parameters encapsulated in Swin V2-B, all patch tokens likely learn similar attention pathways, therefore all visualized methods appear to have widespread attention placed on the input image.  

\section{CMML Dataset Details}
\label{cmml-dataset}
\subsection{Data Collection}
The investigated Chronic myelomonocytic leukemia (CMML) dataset (data statistics shown in Table~\ref{fig:dataset} (a)) was collected from the South Australian Pathology (SA Pathology) laboratory using a Cellavision DI-60 scanner from the period November 2021 to February 2023, in 4 batches.
The blood film staining protocol used a dual Wright’s/Giemsa 0.26\% stain solution and Sorensen’s buffer pH 6.8 from Kinetik.
The scanner detected blood cells on individual images where the cell of interest is centered. 
We used the identified monocytes by the scanner as the raw input images.
The produced monocyte images were then squared or nearly squared in height and width of 352 or 356 pixels, corresponding to a spatial resolution of $36\times36\mu m$.
The collected dataset of images was also manually examined to filter out non-monocytes classified incorrectly by the scanner and images with multiple monocytes. The process resulted in 4,067 monocyte images from 171 individuals. The labels are assigned at the individual level with two classes: Normal and CMML, determined by individual medical records. For each individual included in this study, the number of monocyte images varies from 5 to 171. CMML individuals have on average 46 images vs 17 for normal individuals.
The causes of the variations include: 1) when the WBC count is very low (typically $<0.5\times10^9/\mathrm{L}$), the scanner may have difficulties scanning sufficient WBCs in a study; 
2) normal individuals have fewer monocytes ($<10\%$ of total WBCs) than CMML individuals ($>10\%$); and 3) suspected CMML individuals were repeatedly scanned.
We capped the number of images per individual to 80 which further reduces the samples used for training and testing to 3,899.

\subsection{Motivation}
The CMML dataset depicts a clinically important but difficult diagnostic problem.
In Table~\ref{fig:dataset} (b), we show the result of a human study on 153 monocyte images (53\% are from CMML individuals) rated by 3 hematologists, where the performances are largely inconsistent with the recorded diagnosis from bone marrow biopsy. 
This suggests that individual image-level recognition cannot be done reliably. 
We first make the assumption that the majority of CMML individuals will predominantly have abnormal monocytes, though some monocytes could be normal. 
Then, all the captured image instances of an individual inherit the same label from the individual level, for the purpose of training and testing. 
Finally, in the testing phase, the individual's diagnosis is aggregated by averaging the predictions from the image instances.

From Table~2 in the main manuscript, we show that fitting a vanilla ResNet-50 on this task achieves $90.5\%$ mean accuracy.
With the distribution of approximately $20\%$ individuals belonging to the CMML category and sampled proportionally in the training, validation, and test sets,  this accuracy indicates that the DNN may have found some image cues that correlate to the CMML diagnosis.
The reason for evaluating the CMML dataset is to visualize what image regions have been used to make the model decisions in order to provide insights for the hematologists to understand any morphological/appearance change of CMML in monocytes.

\subsection{Analysis and Discussion}

CAM methods indicate image regions that matter to the model outcome but the region is not meaningful unless we know what is inside the region.
To illustrate, in conventional image classification, we can instantly tell whether a  CAM-highlighted region is part of a dog or other objects, so we can judge whether CAMs make sense.
However, for CMML, we don't have that prior knowledge, therefore we first annotated randomly selected 220 images with nucleus and cytoplasm segmentation by a hematologist. 
These images were used to train a Mask R-CNN~\cite{he2017mask} model to produce predictions for the entire CMML dataset.
With this information and CAPE-produced probabilistic image region contribution, we show that we can summarize the CAPE output of the entire test set to produce a statistical analysis of attention placement on different region types with semantic meanings: nucleus, cytoplasm, and cell exterior region.

The statistical analysis is shown in Table~\ref{tab:semantic_contribution} by summarizing all image predictions (made from the five-fold cross-validation) for the entire dataset.
For each image, we aggregate the image region predictions by horizontal, vertical flipping, and +/- 90\% rotations.
Since an image region can be a square that sits on the boundary of two or more semantic classes, we define six semantic classes shown as the column titles of Table~\ref{tab:semantic_contribution}.
The definition of \textit{simplex} for any semantic class is determined by predefined triplet percentages of (Mask R-CNN) segmentation pixels (Nucleus\%, Cytoplasm\%, Exterior\%) compositing an image region. 
An image region's probabilistic contribution to the overall model decision is assigned to the bag of the closest semantic class determined by the $\mathcal{L}_2$ distance between the image region and the defined semantic class position on the simplex surface.
Finally, for each combination of semantic class and diagnostic class, we compute the mean and standard deviation contribution value of the bag and show that in the corresponding cell in Table~\ref{tab:semantic_contribution}.
We further include the statistics of the CAPE difference between the CMML and Normal classes.
From Table~\ref{tab:semantic_contribution}, we derive several observations such as the following.
\begin{enumerate}
    \item The nucleus region favors the CMML diagnosis more in the (TS) model but stays mutual in the (PT) model. 
    \item The cytoplasm region favors the Normal decision more.
    \item The Cyto-Ext boundary shows a bias towards the Normal class.
    \item The cell exterior region constitutes the largest decision-making and is more biased towards the Normal class. Note that the cell exterior region has the largest image area and hence potentially can host more attention placement.
    \item The rest of the boundaries have relatively less area and do not show a significant bias to either class and hence contribute insignificantly to the overall decision.
    \item The nucleus and cell exterior are the two semantic regions that have the largest standard deviations, meaning they are frequently used to decide the CMML diagnosis.
\end{enumerate}
Therefore, through these observations, one potential research direction is to look into the more fine-grained nucleus morphology analysis and another to examine the potential of red blood cell morphology analysis for CMML. 

\subsection{Dataset Availability}

The ethical approval and data sharing agreement of the CMML research does not cover the public release of the image dataset. Hence the dataset will not be made publicly available.

{
    \small
    \bibliographystyle{ieeenat_fullname}
    \bibliography{supp}
}